\RequirePackage{xcolor}
\documentclass[9pt,shortpaper,twoside,web]{ieeecolor}
\usepackage{generic}
\usepackage{latexsym}
\usepackage{graphicx}
\usepackage{amsfonts,amssymb,amsmath}
\usepackage{hyperref}
\usepackage{cite}
\usepackage{algorithmic}
\usepackage{textcomp}
\usepackage{booktabs}
\usepackage{pgfplots}
\usepackage{multirow}
\usepackage{float}
\usepackage{mathrsfs}
\usepackage[ruled]{algorithm2e}
\usepackage{etoolbox}

\usepackage{soul}
\usepackage{booktabs}
\usepackage{xcolor}
\usepackage{tabularray}
\usepackage{lipsum}
\usepackage{xr}
\usepackage[switch]{lineno}
\usepackage{comment}
\usepackage{subfigure}
\usepackage{hhline}

\modulolinenumbers[5]

\definecolor{RBMEgreen}{HTML}{2F7360}
\makeatletter
\renewcommand{\@cite}[1]{\textcolor{RBMEgreen}{[#1]}}
\makeatother

\definecolor{lei_blue}{HTML}{1767A0}
\definecolor{lei_red}{HTML}{AB3A2A}
\definecolor{lei_orange}{HTML}{D17F2B}
\definecolor{lei_cyanblue}{HTML}{0098B4}
\definecolor{lei_purple}{HTML}{6E6DA2}

\providecommand{\Leireftb}[1]{Table~\ref{#1}}
\providecommand{\Leireffig}[1]{Fig.~\ref{#1}}

\providecommand{\cite}[1]{\cite{#1}}

\def\BibTeX{{\rm B\kern-.05em{\sc i\kern-.025em b}\kern-.08em T\kern-.1667em\lower.7ex\hbox{E}\kern-.125emX}}
\begin{document}
\title{Solving the Inverse Problem of Electrocardiography for Cardiac Digital Twins: A Survey}
\author{Lei Li, Julia Camps, Blanca Rodriguez, and Vicente Grau  
\thanks{Corresponding author: Lei Li (lei.sky.li@soton.ac.uk). 
This work was supported by the CompBioMed 2 Centre of Excellence in Computational Biomedicine (European Commission Horizon 2020 research and innovation programme, grant agreement No. 823712) and CompBiomedX EPSRC-funded grant (EP/X019446/1 to B. Rodriguez).
J. Camps was funded by an Engineering and Physical Sciences Research Council doctoral award.
J. Camps and B. Rodriguez were funded by a Wellcome Trust Fellowship in Basic Biomedical Sciences to Blanca Rodriguez (214290/Z/18/Z).
V. Grau was partially supported by the British Heart Foundation Project under Grant PG/20/21/35082.
}
\thanks{Lei Li is with School of Electronics \& Computer Science, University of Southampton, SO17 1BJ Southampton, UK and
Department of Engineering Science, University of Oxford, OX3 7DQ Oxford, UK.}
\thanks{Vicente Grau is with the Department of Engineering Science, University of Oxford, OX3 7DQ Oxford, UK.}
\thanks{Julia Camps and Blanca Rodriguez are with the Department of Computer Science, University of Oxford, OX1 3QD Oxford, UK.}}


\maketitle
	
\begin{abstract}

Cardiac digital twins (CDTs) are personalized virtual representations used to understand complex cardiac mechanisms. 
A critical component of CDT development is solving the ECG inverse problem, which enables the reconstruction of cardiac sources and the estimation of patient-specific electrophysiology (EP) parameters from surface ECG data. 
Despite challenges from complex cardiac anatomy, noisy ECG data, and the ill-posed nature of the inverse problem, recent advances in computational methods have greatly improved the accuracy and efficiency of ECG inverse inference, strengthening the fidelity of CDTs.
This paper aims to provide a comprehensive review of the methods of solving ECG inverse problem, the validation strategies, the clinical applications, and future perspectives.
For the methodologies, we broadly classify state-of-the-art approaches into two categories: deterministic and probabilistic methods, including both conventional and deep learning-based techniques. 
Integrating physics laws with deep learning models holds promise, but challenges such as capturing dynamic electrophysiology accurately, accessing accurate domain knowledge, and quantifying prediction uncertainty persist. 
Integrating models into clinical workflows while ensuring interpretability and usability for healthcare professionals is essential. 
Overcoming these challenges will drive further research in CDTs.

\end{abstract}

\begin{IEEEkeywords}
\footnotesize Cardiac digital twins, ECG, inverse problem, ECGI, parameter estimation, model personalization, survey.
\end{IEEEkeywords}

\section{Introduction}

\begin{figure*}[t]\center
 \includegraphics[width=0.78\textwidth]{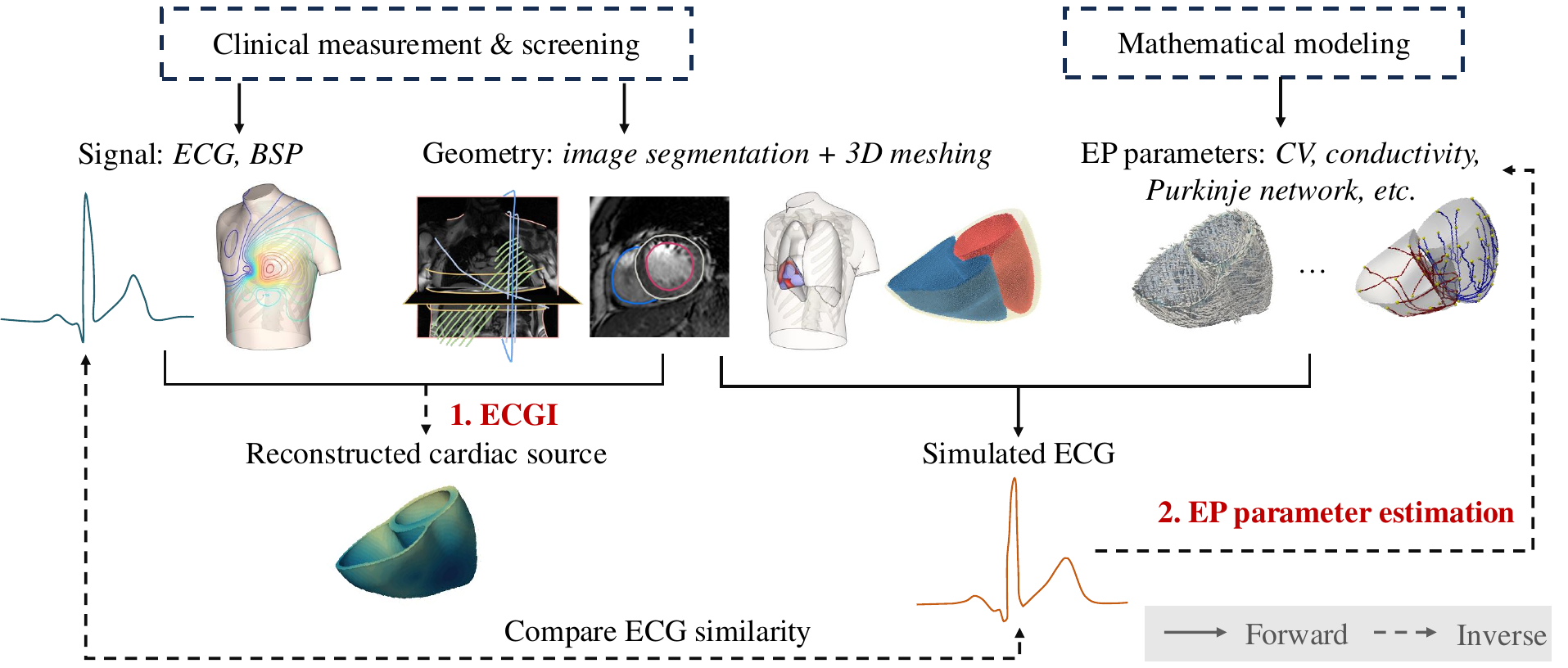}\\[-2ex]
   \caption{Illustration of the electrocardiogram (ECG) inverse problem, including non-invasive cardiac source reconstruction, also known as ECG imaging (ECGI), and electrophysiological (EP) parameter estimation. Here, we take the biventricular modeling as an example. BSP: body surface potential; CV: conduction velocity.
   }
\label{fig:intro:inverseECG}
\end{figure*}

Electrocardiography (ECG) investigates the relationship between the cardiac electrical activity and the resulting potential measured on the torso surface. 
Understanding this relationship is crucial for developing personalized virtual models of cardiac electrophysiology (EP), i.e., cardiac digital twins (CDTs) \cite{journal/MedIA/camps2021,journal/MedIA/gillette2021}. 
Currently, treatment decisions are still primarily based on empirical clinical studies that statistically compare the effects of different treatments across large groups of patients with similar cardiac conditions \cite{journal/NRC/niederer2019}.
Instead, the virtual heart models have shown great promise in providing personalized treatments, such as arrhythmia risk stratification \cite{journal/Nature_C/arevalo2016} and ablation guidance for persistent atrial fibrillation (AF) \cite{journal/Nature_BME/boyle2019}.
This involves using computational models to solve the ECG forward and inverse problems.
The ECG forward problem aims to predict signal on the body surface from known cardiac activity.
Instead, the ECG inverse problem seeks to infer cardiac information from ECG or body surface potential (BSP) \cite{journal/EMBM/macleod1998,journal/MedIA/gillette2021}, as presented in \Leireffig{fig:intro:inverseECG}. 
It can be conducted in two aspects, i.e., the reconstruction of cardiac sources and the estimation of EP parameters.
Both aspects are closely linked with the concept of CDTs \cite{journal/FiP/cluitmans2018,journal/MedIA/gillette2021}.
The reconstruction of cardiac sources provides crucial information about the cardiac electrical activity, which can be integrated into digital twin models to enhance their accuracy in representing the heart \cite{journal/FiP/cluitmans2018}.
On the other hand, the estimation of EP model parameters allows these digital twins to be fine-tuned to reflect individual patient conditions, ensuring that the model behavior closely matches the real cardiac physiological responses \cite{journal/FiP/cluitmans2018}.


However, due to the nature of the ECG inverse problem, the reconstruction of cardiac sources and the estimation of EP parameters share several common challenges.
First, both scenarios involve solving an inherently ill-posed inverse problem, which may not have a unique solution. 
This means that small errors in the input data can lead to significant inaccuracies in the output.
Second, ECG data recorded from the body surface, often using a limited number of electrodes such as the 12-lead ECG with ten electrodes, has limited spatial resolution and is susceptible to noise, artifacts, and inconsistencies.
Third, both scenarios need to account for the complex, inhomogeneous structure of the thorax, including the heart, lungs, and other tissues, which influence how electrical signals propagate to the body surface. 
Accurately modeling these structures poses a significant challenge that affects the reliability of reconstructed cardiac sources and the accuracy of estimated EP parameters.
Compared to the inverse inference of cardiac sources, the inference of EP parameters presents several additional challenges. 
While the cardiac source reconstruction requires computational resources, it is generally not considered excessively time-consuming. 
In contrast, EP parameter estimation is far more computationally intensive due to its reliance on sophisticated anatomical modeling, iterative simulations, and complex parameter fitting, particularly when using high-resolution models.
Also, EP parameters can be interdependent, meaning that changes to one parameter may affect others, which further complicates the optimization and fitting process \cite{journal/CBM/costa2022}.
Unlike standard ECG imaging, which is directly validated against real-time measurements, the EP parameter estimation requires comprehensive validation to ensure the generated CDTs accurately represent the unique anatomy and EP of the patient.

\subsection{History: From Electrocardiography to Cardiac Digital Twins} 

Investigations of cardiac electrophysiology can be traced back to 1842 when Carlo Matteucci identified signs of cardiac electrical activity in frog hearts \cite{journal/ACP/matteucci1842}. 
Dr. Augustus Waller furthered this research by recording the first human electrocardiogram in 1887, using a capillary electrometer and electrodes placed on the chest and back of a human subject \cite{journal/JP/waller1887}. 
His study demonstrated that electrical activity preceded ventricular contraction. 
Subsequently, in 1893, Einthoven coined the term ``electrocardiogram" to describe the cardiac waveforms \cite{journal/PAEJP/einthoven1895}.
The American Heart Association (AHA) standardized the 12-lead ECG in 1954 \cite{journal/Cir/wilson1954}, paving the way for advancements in cardiac electrophysiology. 
In 1960, Denis Noble developed one of the first mathematical models of a cardiac electric cell based on Hodgkin-Huxley equations, further refining it to characterize the long-lasting action and pacemaker potentials observed in the Purkinje fibres of the heart \cite{journal/nature/noble1960,journal/JP/hodgkin1952,journal/JP/noble1962}. 
This laid the groundwork for cardiac simulations using computer models.
Meanwhile, BSP mapping emerged as an alternative method for non-invasive and high-resolution measurement of human cardiac electrical activity in 1963 \cite{journal/Cir/taccardi1963}. 
In 1970, Durrer \textit{et al.} \cite{journal/Cir/durrer1970} conducted a seminal investigation into the electrical behavior of the human heart at the organ level.
Since 1972, numerous studies have tackled the electrocardiography inverse problem using analytical and computational models \cite{journal/TBME/martin1972,journal/TBME/martin1975,journal/TMI/barr1977,journal/AC/franzone1978,journal/TBME/yamashita1984}, experimental animals \cite{journal/CR/barr1978, journal/CR/messinger1990}, and human subjects \cite{journal/Nature_medicine/ramanathan2004,journal/CAE/sapp2012}. 
The formalization of inverse procedures reconstructing bioelectric sources from BSP mapping (BSPM) led to the development of non-invasive electrocardiographic imaging (ECGI), a term formally coined by Yoram Rudy's lab in 1997 \cite{journal/Cir/oster1997}.
While ECGI can be patient-specific in terms of geometry and electrical potentials, it does not typically include a full-scale, personalized model of the cardiac EP properties.

\begin{table} [t] \center
    \caption{Search engines and expressions used to identify potential papers for review. 
     }
\label{tb:intro:search item}
{\small
\begin{tabular}{p{0.75cm}p{7.2cm}}
\hline
Engine                 & Google Scholar, PubMed, IEEE-Xplore, and Citeseer\\
\hline
\multirow{12}{*}{Term}  &``Electrocardio*" or ``ECG" or ``EKG" or ``electrophysiolog*" or ``EP" or ``body surface potential" or ``BSP" \textbf{and} \\ \cline{2-2}
                       & ``Inverse problem/ issue" or ``inverse solution/ estimat*/ infer*/ reconst*" or ``electrocardiographic imaging/ mapping"  or ``ECG-imaging" or ``ECGI" or ``activation imaging" or ``noninvasive imaging"  or ``personaliz*/ calibrat*" \textbf{and} \\ \cline{2-2} 
                       & ``Electrical activity/ excitation" or ``heart/ epicardial/ endocardial /atrial / transmembrane potential" or ``activation sequence/ wavefront/ time" or ``pacing site" or ``parameter" or ``onset location/ root note/ earliest activation site/ Purkinje" or ``conducti*" or ``origin of cardiac activation" or ``excitability"\\ 
\hline
\end{tabular} }\\
\end{table}

In constrast, CDT is a comprehensive, personalized model that incorporates both the electrical and structural properties of the heart, including detailed EP parameters.
The concept of the digital twin was formally introduced in 2002 by Michael Grieves, though it can be traced back to NASA's use for the Apollo 13 rescue mission in 1970 \cite{journal/MTA/guo2022}. 
The Physiome Project was initiated in 2003 with the aim of establishing a framework for modeling the human body, utilizing computational methods to integrate biochemical, biophysical, and anatomical information across cells, tissues, and organs \cite{journal/NRMCB/hunter2003}. 
Starting from 2014, many companies like Dassault, Siemens, and ANSYS embraced the term ``digital twins" in their marketing initiatives. 
For example, Dassault Systèmes initiated the Living Heart Project in 2014, which is the first comprehensive computer model of the human heart, integrating many functional aspects, i.e., blood flow dynamics, mechanical properties, and electrical conduction \cite{journal/JPM/armeni2022}.
In recent years, digital twin tools have become popular in healthcare, such as CDTs \cite{journal/EHJ/corral2020}, brain digital twins \cite{journal/VI/li2023}, etc.
In 2016, the US Food \& Drug Administration (FDA) Center for Devices and Radiological Health released the first guidance to enable the use of modeling and simulation to develop in silico trials \cite{journal/Methods/viceconti2021}.
Notably, Trayanova's lab received FDA approval in 2019 for a randomized clinical trial involving 160 patients, termed OPTIMA (OPtimal Target Identification via Modeling of Arrhythmogenesis), to demonstrate the effectiveness of CDTs in guiding atrial ablation procedures \cite{journal/Nature_BME/boyle2019}.


\begin{table*}[t]
    \centering  
    \caption{Summary of the related representative existing surveys. DL: deep learning.}
    \label{tb:intro:review paper}
    \resizebox{0.75\textwidth}{!}{	
    \begin{tabular}{lll@{\ \,}l}
        \hline
        Source \& Publish Year & Scope & Gap \\
        \hline
        Corral \textit{et al.} (2020) \cite{journal/EHJ/corral2020} & cardiac digital twins & challenges and clinical applications  \\
        Coorey \textit{et al.} (2022) \cite{journal/npj/coorey2022} & cardiac digital twins & challenges and clinical applications  \\
        Bear \textit{et al.} (2015) \cite{journal/CAE/bear2015}           & ECG forward problem         & not for inverse problem \\ 
        Bergquist \textit{et al.} (2021) \cite{journal/CBM/bergquist2021} & ECG forward problem         & not for inverse problem \\ 
        Gulrajani \textit{et al.} (1998) \cite{journal/EMBM/gulrajani1998} & ECG forward and inverse problem & conventional solutions \\   
        Uhlmann \textit{et al.} (2014) \cite{journal/BMS/uhlmann2014}     & inverse problem             & not for ECG; conventional solutions \\
        Calvetti \textit{et al.} (2018) \cite{journal/WIRCS/calvetti2018} & inverse problem             & not for ECG; conventional solutions \\ 
        Genzel \textit{et al.} (2022) \cite{journal/TPAMI/genzel2022}     & inverse problem             & not for ECG; mainly DL-based solutions \\ 
        Siontis \textit{et al.} (2021) \cite{journal/Nature_RC/siontis2021}& DL-based ECG analysis      & not for inverse problem \\
        Somani \textit{et al.} (2021) \cite{journal/EP/somani2021}        & DL-based ECG analysis       & not for inverse problem \\
        Bifulco \textit{et al.} (2021) \cite{journal/Heart/bifulco2021}   & ECG inverse problem         & applications of computational modeling \\  
        Macleod \textit{et al.} (1998) \cite{journal/EMBM/macleod1998}    & ECG inverse problem         & conventional ECGI solutions \\
        Cluitmans \textit{et al.} (2018) \cite{journal/FiP/cluitmans2018} & ECG inverse problem         & conventional ECGI solutions and applications \\
        Hernandez \textit{et al.} (2023) \cite{journal/MBEC/hernandez2023}& ECG inverse problem         & atrial ECGI \\
        Loewe \textit{et al.} (2022) \cite{book/ITSCE/loewe2022}          & ECG inverse problem         & partially EP parameter estimation \\
        \hline
    \end{tabular}
    }
\end{table*}

\subsection{Study Inclusion and Literature Search}
In this work, we aim to provide readers with a survey of the state-of-the-art solutions to ECG inverse problems.
There exist several modalities for cardiac electrical activity assessment, such as ECG, BSP, electrogram (EGM), electro-anatomic mapping (EAM), vectorcardiography (VCG).
However, in the context of solving the ECG inverse problem, we usually only involve non-invasive ECG and BSP as the observed data.
Note that compared to standard 12-lead ECG, BSPM is, however, not yet widely available as a diagnostic modality \cite{journal/MedIA/zettinig2014}.
For cardiac activity representation, three major source models are employed, i.e., surface potential models, activation models, and transmembrane potential (TMP).
For model personalization, the EP parameters to be estimated can be conductivities, conduction velocities (CVs), earliest activation sites (EASs), tissue excitability, etc.

To ensure comprehensive coverage, we have screened publications mainly from the last 24 years (2000-2024) related to this topic.
Our main sources of reference were Internet searches using engines, including Google Scholar, PubMed, IEEE-Xplore, Connected Papers, and Citeseer.
To cover as many related studies as possible, flexible search terms have been employed when using these search engines, as summarized in \Leireftb{tb:intro:search item}.
There are several research teams and industry companies at the forefront of computational cardiology and electrocardiography, working to advance our understanding of cardiac function, arrhythmias, and heart-related disorders through computational methods and simulations \footnote{Continuously updating compilation of cardiac digital twin resources: \url{github.com/lileitech/Awesome-Cardiac-Digital-Twins}}.
We followed their recent works to obtain state-of-the-art methodologies and clinical applications.
Both peer-reviewed journal and conference papers were included here.
In the way described above, we have collected a comprehensive library of more than 200 papers.
Note that we normally only picked the most detailed and representative ones for this review when we encountered several papers from the same authors about the same subject.

\subsection{Existing Survey from Literature}

\Leireftb{tb:intro:review paper} lists current review papers related to the topic, i.e., ECG inverse problem.
One can see that the scopes of current review works are different from ours, though with partial overlaps.
For example, several surveys have been conducted to summarize the current challenges and existing or potential clinical applications of ECGI \cite{journal/FiP/cluitmans2018} or CDTs \cite{journal/EHJ/corral2020,journal/npj/coorey2022}.
Furthermore, there are several existing reviews focusing on summarizing solutions of inverse problems but not for ECG data \cite{journal/BMS/uhlmann2014,journal/WIRCS/calvetti2018,journal/TPAMI/genzel2022}.
Alternatively, there exist few surveys on the ECG analysis, but not focusing on the inverse problem \cite{journal/Nature_RC/siontis2021,journal/EP/somani2021}.
Bifulco \textit{et al.} \cite{journal/Heart/bifulco2021} provided the potential clinical applications of the ECG inverse inference strategies, but it is not a methodological survey.
Although Macleod \textit{et al.} \cite{journal/EMBM/macleod1998}, Cluitmans \textit{et al.} \cite{journal/NHJ/cluitmans2015} and Hernandez \textit{et al.} \cite{journal/MBEC/hernandez2023} performed a methodology survey on the ECG inverse problem, they only summarized the ECG inverse inference studies for cardiac source reconstruction, i.e., ECGI.
Loewe \textit{et al.} \cite{book/ITSCE/loewe2022} partially reviewed the ECG inverse inference works on EP parameter estimation in terms of cellular and tissue levels.
Here, we provide a comprehensive and up-to-date review of ECG inverse inference in a broader range, encompassing both cardiac source reconstruction and EP parameter estimation. 
The communities engaged in these two areas of research do not completely overlap and sometimes even use different terminologies. 
For example, in the narrow definition of the inverse-ECG problem, i.e., ECGI, the forward model typically refers to the relationship between cardiac sources and ECG data. 
In contrast, within the model personalization community, the forward model often pertains to the forward EP model. 
This survey aims to bridge the gap between these communities, fostering conversation and collaboration.

\begin{figure*}[t]\center
 \includegraphics[width=0.84\textwidth]{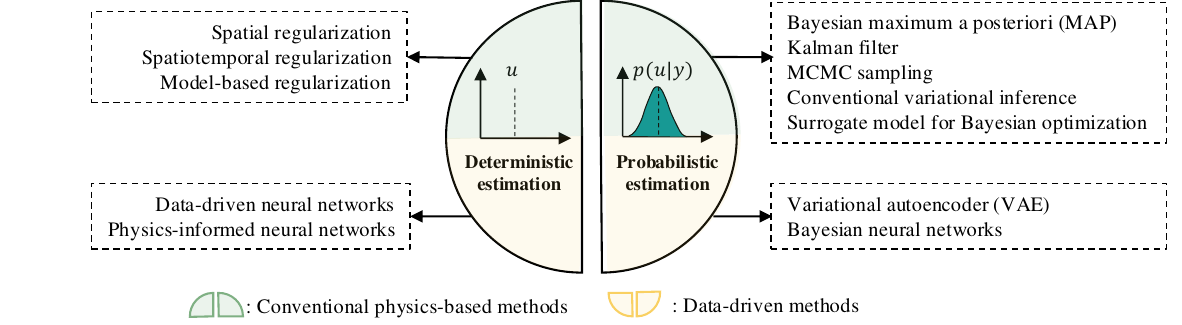}\\[-2ex]
   \caption{Summary of the ECG inverse inference methodologies. MCMC: Markov Chain Monte Carlo.
   }
\label{fig:method:summary}
\end{figure*}

\section{Methodology} \label{method}

\subsection{Definition of the Problem}

The ECG inverse problem involves inferring internal cardiac information from surface ECG measurements. 
This problem can be formulated in two ways:
\begin{enumerate}
    \item \textbf{Reconstruction of cardiac sources (ECGI):} It aims to reconstruct the cardiac electrical activity from the recorded ECG or body surface potentials (BSPs) for a non-invasive visualization of cardiac activity. 
    The relationship between the ECG data $\boldsymbol{y}(\boldsymbol{s}, t)$ and the cardiac sources $\boldsymbol{u}(\boldsymbol{s}, t)$ at position $\boldsymbol{s}$ and time instant $t$ is typically modeled as a linear system:
    \begin{equation}
        \boldsymbol{y}(\boldsymbol{s}, t) = \boldsymbol{H} \boldsymbol{u}(\boldsymbol{s}, t) + \epsilon,
    \end{equation}
    where $\boldsymbol{H}$ is approximated as a forward matrix, and $\epsilon$ accounts for noise. 
    The goal is to estimate $\boldsymbol{u}(\boldsymbol{s}, t)$ by minimizing the fitting of the observational data $\boldsymbol{y}(\boldsymbol{s}, t)$: 
    \begin{equation}
        \min _{\boldsymbol{u}(\boldsymbol{s}, t)} \| \boldsymbol{y}(\boldsymbol{s}, t) - \boldsymbol{H}\boldsymbol{u}(\boldsymbol{s}, t) \|_2^2.
    \end{equation}  
    Note that this linear model specifically applies to potential-based ECG inverse problems. 
    Activation-based models, which often involve non-linear relationships, are not directly addressed by this formulation \cite{journal/TMI/liu2006,journal/TMI/yang2018,journal/TBME/schuler2021}.
    \item \textbf{Estimation of EP model parameters:} It focuses on personalize cardiac models by estimating patient-specific EP parameters within a mathematical model of cardiac electrophysiology, such as the Aliev-Panfilov (AP) model \cite{journal/CSF/aliev1996} or Eikonal model \cite{journal/MedIA/camps2021}.
    The AP model can be defined as,
    \begin{equation}
        \begin{aligned}
        \frac{\partial u}{\partial t} &= \nabla \cdot (\boldsymbol{D} \nabla \boldsymbol{D}) - cu(u - \theta)(u - 1) - \frac{\partial u}{\partial z}, \\
        \frac{\partial z}{\partial t} &= \epsilon(u, z) \left(-z - cu(u - \theta - 1)\right),
        \end{aligned}
    \end{equation}
    where \( u \) is the normalized transmembrane action potential, \( z \) is the recovery current, \(\boldsymbol{D}\) is the diffusion tensor, \(c\) controls the repolarization, and \( \theta\) is tissue excitability parameter.
    The Eikonal equation describes the propagation of the activation wavefront:
    \begin{equation} 
        \begin{cases}
        |\nabla \tau(\boldsymbol{s})| = \dfrac{1}{v(\boldsymbol{s})}, & \boldsymbol{s} \in \Omega, \\
        \tau(\boldsymbol{s}) = \tau_0(\boldsymbol{s}), & \boldsymbol{s} \in \Gamma_0,
        \end{cases} \label{eq:eikonal}
    \end{equation}
    where \( \tau(\boldsymbol{s}) \) represents the activation time at point \( \boldsymbol{s} \) within the cardiac tissue domain \( \Omega \), while \( v(\boldsymbol{s}) \) denotes the local CV. 
    The boundary conditions are defined by the set of EASs \( \Gamma_0 \), with \( \tau_0(\boldsymbol{s}) \) specifying the activation times at these sites.
    Taking the Eikonal model as an example, its parameter optimization objective function can be formulated as,
    \begin{equation}
        \min_{\boldsymbol{v}, \tau_0} \| \boldsymbol{y}_{\text{ECG}} - \boldsymbol{y}_{\text{model}}(\boldsymbol{v}, \tau_0) \|_2^2 + \lambda \mathcal{R}(\boldsymbol{v}),
    \end{equation}
    where $\boldsymbol{y}_{\text{ECG}}$ is the measured signal, while $\boldsymbol{y}_{\text{model}}(\boldsymbol{v}, \tau_0)$ represents the simulated ECG signals using the forward EP model with CVs $\boldsymbol{v}$ and EASs $\tau_0$; 
    $\mathcal{R}(\boldsymbol{v})$ imposes certain constraints on the estimated parameters, and $\lambda$ is a regularization term.
\end{enumerate}
Both are essential for creating CDTs, but they may involve different methods.
The state-of-the-art approaches can be coarsely separated into two kinds: deterministic and probabilistic methods for both \cite{book/SSBM/kaipio2006}, as presented in \Leireffig{fig:method:summary}.
Deterministic approaches in cardiac electrophysiology involve minimizing a cost function that quantifies the discrepancy between the observed data and the model predictions.
Rather than aiming solely to minimize a cost function, probabilistic models embrace uncertainty and provide a probabilistic distribution over possible outcomes. 
We will review these two approaches separately and summarize their application to different formulations of ECG inverse inference. 
While we categorize the methodologies into deterministic and probabilistic, it is important to note that these categories often overlap, which will be emphasized when introducing these methods. 

\subsection{Deterministic Estimation Approaches} 



\subsubsection{Spatial Regularization} 

To stabilize and improve the solution of the ECG inverse problem, spatial regularization methods are widely employed for ECGI \cite{journal/FiP/figuera2016}. 
These methods introduce spatial constraints to promote smoothness in the estimated electrical activities across the heart surface at different orders of derivatives \cite{journal/TMI/rodrigo2017}. 
The commonly used spatial regularization methods include Tikhonov regularization \cite{journal/Nature_medicine/ramanathan2004}, truncated singular value decomposition (TSVD) \cite{conf/ISHIB/sarikaya2010}, truncated total least squares (TTLS) \cite{journal/TBME/shou2008}, L1-norm regularization \cite{journal/ABME/ghosh2009}, and total variation (TV) regularization \cite{journal/TMI/xu2014}. 
These approaches aim to minimize the impact of noise and improve the accuracy of the inverse solution.
For example, the Tikhonov regularization, including both zero-order and first-order variants, is the most widely used regularization technique \cite{journal/Nature_medicine/ramanathan2004,journal/TBME/wang2009a,journal/FiP/kara2019,journal/TBME/bear2020,journal/TBME/schuler2021,journal/CBM/rababah2021}. 
It minimizes the mean squared error from the body-heart transformation while simultaneously penalizing the L2 norm of the inverse heart surface potential (HSP) solution. 
The objective function can be formulated as,
\begin{equation}
    \min _{\boldsymbol{u}(\boldsymbol{s}, t)}\left\{\|\boldsymbol{y}(\boldsymbol{s}, t)-\boldsymbol{H} \boldsymbol{u}(\boldsymbol{s}, t)\|_2^2+\lambda_{Tikh}^2\|\Gamma \boldsymbol{u}(\boldsymbol{s}, t)\|_2^2\right\},  \label{eq:Tikhonov}
\end{equation}
where the notation $\| \cdot \|^2$ represents the L2 norm, $\lambda_{Tikh}$ denotes the regularization coefficient, and $\Gamma$ is the operator constraining the HSP solution $u(\mathbf{s}, t)$, which is an identity matrix for zero-order regularization and a spatial gradient operator for first-order regularization to increase the spatial smoothness of the inverse solution.
This dual process effectively suppresses the unreliable components in the solution and enhances overall smoothness for improved accuracy \cite{journal/sensors/wang2023}.
The choice of \(\lambda_{\text{Tikh}}\) is crucial for enhancing reconstruction quality, and the L-curve method is commonly employed to determine the optimal value \cite{conf/FIMH/chamorro2017,journal/sensors/wang2023}.
TSVD regularization method directly filters small singular values of the transfer matrix $\boldsymbol{H}$ and thus be robust to noises.
TTLS regularization is effective in processing geometric errors that exist in the ECG inverse problem \cite{journal/TBME/shou2008}.
L1-norm and TV regularization can tackle specific challenges in the inverse problems, such as noise suppression, sparsity, and edge preservation \cite{journal/JCE/zhou2018}. 
Note that although these methods have been extensively applied to HSP based solutions, they can be applied to other linear inverse problem representation, such as TMPs \cite{journal/TBME/melgarejo2022}.

\subsubsection{Spatiotemporal Regularization} 

While spatial regularization methods are effective and widely used in ECGI, they often overlook the temporal dynamics of cardiac electrical activity. 
To fuse the space-time correlations into the ECGI, spatiotemporal regularizations are therefore employed for robust estimation of cardiac sources.
For instance, Twomey regularization was proposed as a modification of Tikhonov regularization to incorporates temporal a priori information \cite{journal/TBME/oster1992,journal/TBME/ghodrati2006}.
It adjusts the second term in Eq. \ref{eq:Tikhonov} to minimize the difference between the solution and the prior estimate from the previous time instant.
Alternative, one could introduce several additional constrains, including temporal constraints, to increase robustness to changes in the values of the regularization parameters \cite{journal/TBME/brooks1999,journal/TBME/messnarz2004,journal/MBEC/serinagaoglu2013}.
Yao \textit{et al.} \cite{journal/SR/yao2016} proposed a spatiotemporal regularization method by regularizing both the spatial and temporal smoothness, with the following objective function,
\begin{equation}
\begin{aligned}
& \min_{\boldsymbol{u}(s, t)} \sum_t\bigg\{ \|\boldsymbol{y}(\boldsymbol{s}, t) - \boldsymbol{H} \boldsymbol{u}(\boldsymbol{s}, t)\|_2^2 + \lambda_s^2\|\Gamma \boldsymbol{u}(\boldsymbol{s}, t)\|_2^2 \\
& + \lambda_t^2 \sum_{\tau=t-\omega / 2}^{\tau=t+\omega / 2}\|\boldsymbol{u}(\boldsymbol{s}, t) - \boldsymbol{u}(\boldsymbol{s}, \tau)\|_2^2 \bigg\},
\end{aligned}
\end{equation}
where $\lambda_s$ and $\lambda_t$ are the spatial and temporal regularization terms, respectively, and $\omega$ denotes the selected time window to incorporate the temporal correlation.
Recently, frequency-guided model have be widely applied for imaging the spatiotemporal behavior of the heart \cite{journal/TMI/liu2006,journal/TBME/zhou2016,journal/TBME/coll2017,journal/TMI/yang2018}.
Alternatively, splines-based methods can use spline functions to represent the electrical activity on the heart surface over time. 
They take advantage of the smoothness properties of splines to regularize the inverse solution, effectively capturing both spatial and temporal variations in the electrical potentials \cite{journal/MBEC/onak2019}.
Erem \textit{et al.} \cite{journal/TMI/erem2013,journal/TMI/erem2014} employed the temporal spline model to reconstruct TMP and ATM and localize pacing sites from BSP. 
However, their approach faced the challenge of determining the optimal number of spline functions in advance. 
To overcome this limitation, more flexible adaptive splines have been proposed to reconstruct cardiac sources, enabling data-driven selection of spline functions from a large set of basis elements through a stepwise process \cite{journal/MBEC/onak2019, journal/TBME/onak2021}.
Recently, to achieve regularization over space and time simultaneously, Cluitmans \textit{et al.} \cite{journal/MBEC/cluitmans2018} applied regularization in the sparse wavelet domain for reconstruct activation wavefronts and epicardial potential (EpiP), namely wavelet-promoted spatiotemporal regularization.
Note that several studies have employed spatiotemporal Bayesian maximum a posteriori (MAP) or Kalman filters for ECGI \cite{conf/WCMPBE/aydin2009,conf/ECIFMBE/jiang2009,journal/TBME/liu2010}, which however belong to probabilistic estimation and will be discussed further in Sec. \ref{method:PEA:MAP} and \ref{method:PEA:kalman}.

\begin{figure*}[t]\center
 \includegraphics[width=0.7\textwidth]{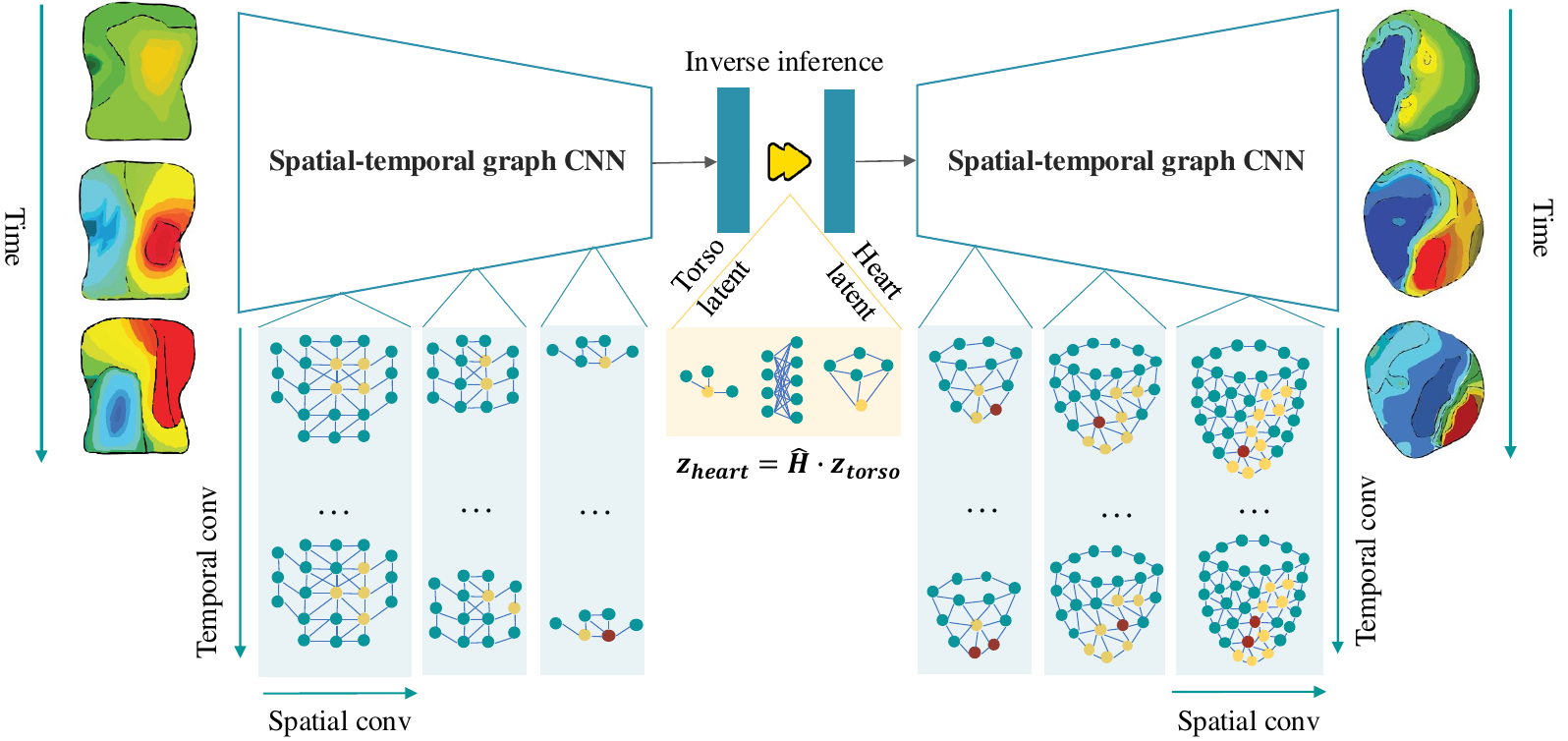}\\[-2ex]
   \caption{Example of the physics-based model based on graph convolutional neural networks (CNNs) for the inverse inference of heart surface potential from body surface potential (BSP) \cite{journal/TMI/jiang2022}. Illustrations designed referring to Jiang \textit{et al.} \cite{journal/TMI/jiang2022} and Bear \textit{et al.} \cite{journal/CAE/bear2018}. 
   }
\label{fig:method:DL_physics_net}
\end{figure*}

\subsubsection{Model-based Regularization} 

Most existing regularization methods primarily rely on the optimization of the given observations, which are normally corrupted by noise in practice.
In contrast, model-based regularization can integrate prior physiological knowledge about the electrical propagation inside the heart to constrain the solution space \cite{journal/MBEC/cluitmans2017}. 
Compared to the above regularization methods that primarily focus on ECGI, model-based methods have been widely applied to both ECGI and EP parameter estimation.

For ECGI, model-based regularization can enhance the representation of cardiac electrical activity by incorporating detailed models of activation and wave propagation.
For instance, step jump functions and logistic functions have been employed to model the activation of action potential \cite{journal/ABME/pullan2001,journal/ABME/van2009}.
Parameterized curves have been utilized to represent the wavefront velocity as trigonometric functions, while the potential has been modelled using the step response of a second-order linear system \cite{journal/TBME/ghodrati2006}.
The ECGI problems can also be formulated via level-sets by evolving a boundary from an initial region to minimize a filtered residual error, allowing for iterative refinement of the solution.
The major advantage of level-set based methods is that no isotropy assumptions are required, allowing for more flexible modeling and accurate representation of complex geometries and phenomena within the heart \cite{conf/FIMH/calderero2005,conf/FIMH/chavez2015}.
Cluitmans \textit{et al.} \cite{journal/MBEC/cluitmans2017} introduced a electrophysiology based regularization to model the cardiac surface potential using a set of basis vectors. 
The epicardial potential is then reconstructed as a sparse combination of these vectors, based on EP principles.
To estimate transmural potentials throughout the myocardium, 3D biophysical EP simulation models have been employed to offer spatiotemporal constraints of the inverse issue \cite{journal/TBME/he2003,journal/TMI/nielsen2013,journal/TBME/wang2009b}.
Similar strategies have been previously employed for predicting onset activation location \cite{journal/TBME/li2001} and reconstrucing 3D activation time from BSP \cite{journal/PMB/he2002,journal/TMI/liu2006,journal/TMI/han2008}.
Furthermore, there exist a few model-based ECGI studies that optimize the model parameters via maximizing the correlation between simulated and measure BSP, which sometimes could be quite computational expensive \cite{journal/JE/van2013,journal/FiP/potyagaylo2019}.

For EP parameter estimation, model-based regularization serves a different purpose. 
It aims to identify patient-specific parameters rather than applying generalized physiological models that are commonly used for ECGI \cite{journal/TBME/wang2009b}.
This is generally achieved by optimization methods based on gradient descent \cite{journal/JNMBE/grandits2021,journal/TBME/grandits2023}.
Furthermore, Chinchapatnam \textit{et al.} \cite{journal/TMI/chinchapatnam2008} proposed an adaptive zonal decomposition iterative algorithm to estimate the local CV and conductivity within the Eikonal model.
Similarly, Sermesant \textit{et al.} \cite{journal/MedIA/sermesant2012} defined the regularization term as the energy of a simplified dynamical electromechanical model of the heart to estimate EASs and local CV.

\subsubsection{Data-Driven Neural Networks}

The notable increase in computational resources and available data have facilitated the development of data-driven models for the ECG inverse inference.

For ECGI, traditional methods often rely on regularization techniques and/ or biophysical models to stabilize the solution, but these approaches may struggle with accuracy and robustness in the presence of data noise or model errors.
In contrast, data-driven neural networks can learn complex mappings between BSP and cardiac source directly from data, without the need for explicit biophysical modeling \cite{journal/EP/bacoyannis2021}. 
They do not rely on strong assumptions about the underlying physiology, making them suitable for cases where the true underlying model is not well-known, varies significantly across patients, or exhibits unexpected abnormal behavior.
Therefore, they offer a promising alternative for solving the ECG inverse problem, particularly in scenarios with high variability or uncertainty in patient-specific cardiac physiology.
The pioneering data-driven ECGI work was conducted by Yang \textit{et al.} \cite{journal/TBME/yang2017}, who utilized the 12-lead ECG and convolutional neural networks (CNNs) to localize the origins of premature ventricular contractions (PVCs). 
They developed two distinct CNN models: one to classify the PVC origin across different segments and another to distinguish between endocardial (Endo) and epicardial (Epi) origins.
Later, Meister \textit{et al.} \cite{conf/STACOM/meister2021} designed a deep learning method based on graph CNNs to estimate the depolarization patterns in the myocardium with scars. 
Similary, Mu \textit{et al.} \cite{conf/MICCAI/mu2021} combined graph CNN and iterative soft threshold algorithm to reconstrcuct TMP and ATM from BSP.
By leveraging the inherent graph structure of the cardiac anatomy, graph CNNs enable the extraction of spatial dependencies and patterns, facilitating a more nuanced representation of ECG data. 
Tenderini \textit{et al.} \cite{journal/SIAM/tenderini2022} utilized a PDE-aware deep learning model for EpiP reconstruction from ECG data.
Camara \textit{et al.} \cite{journal/FiP/camara2021} converted the BSP into an image represented by a 3D matrix and then employed CNNs to predict the locations of atrial fibrillation drivers from BSP.
Cheng \textit{et al.} \cite{conf/EMBC/cheng2021} employed fast iterative shrinkage/thresholding network to reconstruct TMP from ECG.
Chen \textit{et al.} \cite{journal/Sensors/chen2022} tested the feasibility of several different neural networks, including CNN, fully connected neural network, and long short-term memory, for the reconstruction of EpiP and ATM as well as the localization of pacing sites.
Furthermore, time delay neural networks have been employed to extract both current and previous timestep values of BSP to predict the HSP \cite{conf/FIMH/karoui2019,journal/FiP/karoui2021,conf/EMBC/malik2018}.
Interestingly, Karoui \textit{et al.} \cite{journal/FiP/karoui2021} found that direct data-driven ATM reconstruction outperformed the two-stage approach, which first reconstructs HSP via a neural network and then estimates the ATM from the HSP using intrinsic deflection time. 
This is because small fluctuations in the reconstructed HSP can lead to errors in the subsequent computation of ATM.

For EP parameter estimation, the first deep learning based EP parameter personalization model was proposed by Neumann \textit{et al.} \cite{conf/MICCAI/neumann2015}, which however belongs to probabilistic estimation approaches (see Sec. \ref{method:PEA}).
This is because they reformulated the EP parameter estimation problem into a Markov decision process and applied reinforcement learning to personalize the global CVs and Purkinje network \cite{journal/MedIA/neumann2016}.
Later, Giffard \textit{et al.} \cite{journal/TBME/giffard2018} employed transfer learning to estimate local CVs and pacing sites from BSP for personalized cardiac resynchronization therapy (CRT).
Nevertheless, these purely data-driven models generally lack interpretability and may underperform in scenarios with limited data due to their data-intensive training requirements.

\subsubsection{Physics-Informed Neural Networks}

Physics-informed neural networks (PINNs) are data-efficient as they leverage both available data and the known physics equations \cite{journal/NRP/karniadakis2021}.
They have recently been used to solve inverse problems with hidden physics, including the ill-posed ECG inverse problems \cite{journal/TASE/xie2023}.
This can be achieved either via introducing underlying physical constraints as an additional loss \cite{journal/TASE/xie2023,conf/MICCAI/ye2023}, or designing specialized network architectures that satisfy the physics laws \cite{journal/TMI/jiang2022,journal/FCM/herrero2022}.

For ECGI, Xie \textit{et al.} \cite{journal/TASE/xie2023} combined the data-driven loss and physics-based loss to inversely predict the HSP from BSP and evaluated their framework in a 3D torso-heart geometry.
Kashtanova \textit{et al.} \cite{conf/STACOM/kashtanova2022} proposed a APHYN-EP model, which decomposing the transmembrane dynamics into a physical term and a data-driven term, respectively.
The data-driven deep learning component is tailored to capture information that the incomplete physical model fails to represent. 
This concept was later formally defined as the hybrid neural EP model by Jiang \textit{et al.} \cite{journal/TMI/jiang2024}.
Instead of introducing additional loss, Jiang \textit{et al.} \cite{journal/TMI/jiang2022} proposed a spatial-temporal graph CNN-based model, where the geometry-dependent physics between BSP and HSP can be explicitly modelled via a bipartite graph over their graphical embeddings, as presented in \Leireffig{fig:method:DL_physics_net}.

For EP parameter estimation, Ye \textit{et al.} \cite{conf/MICCAI/ye2023} proposed a spatial-temporally adaptive PINN framework for both ECG simulation and AP model parameter estimation.
Herrero \textit{et al.} \cite{journal/FCM/herrero2022} developed a PINN model for action potential reconstruction and EP parameter estimation from sparse amounts of EP data.
Instead, Sahli \textit{et al.} \cite{journal/FiP/sahli2020} used PINNs in the forward model to interpolate activation time maps (ATMs) and estimate CV maps from a few sparse activation measurements.
They incorporated a physics-informed regularization loss prescribed by the Eikonal equation, which describes the relationship between ATMs and the spatial gradient of CV.
Although they did not inversely infer ATMs from BSP or ECG, a similar idea can be employed to achieve the ECG inverse inference.
In general, PINNs for ECG inverse inference are mainly used for 1D/2D in silico data, as they face significant challenges, such as hyperparameter tuning, managing complex geometries, and spectral bias. 
These issues become more pronounced in realistic 2D/3D settings, due to their complex geometries and need for accurate spatial derivatives, leading to high computational requirements \cite{journal/FCM/herrero2022,conf/MICCAI/ye2023}.


\subsection{Probabilistic Estimation Approaches} \label{method:PEA}


Unlike deterministic techniques, probabilistic estimation methods provide solutions in the form of probability distributions. 
Specifically, these approaches can naturally incorporate spatiotemporal information by using spatiotemporal priors or modeling state transitions \cite{journal/MBEC/erenler2019}.
In fact, deterministic methods such as Tikhonov regularization can be formulated as a special and simplified case of Bayesian MAP estimation under linear models with Gaussian priors \cite{journal/MBEC/erenler2019}. 
In these cases, the regularization parameter effectively represents prior knowledge about the model.
Probabilistic methods extend this by not only incorporating such prior information but also by providing a distribution of possible solutions. 
This distribution allows for comprehensive statistical analysis and helps address uncertainties inherent in the data and models relevant to ECG inverse inference \cite{journal/TBME/martin1975}.

\subsubsection{Bayesian MAP Estimation} \label{method:PEA:MAP}

In Bayesian MAP estimation, the solution is selected by maximizing the posterior probability of the sources given the measurements. 
For ECGI, this maximization aims to find the most likely cardiac source distribution that matches both prior physiological knowledge and observed ECG measurements, given a specific forward model \cite{journal/TBME/van1999}.
Greensite \textit{et al.} \cite{journal/TBME/greensite2003} proposed using spatial and spatiotemporal Bayesian MAP methods for bioelectric (including ECG) inverse problems.
Similarly, Zhou \textit{et al.} \cite{journal/TBME/zhou2018b} used a spatiotemporal Bayesian MAP approach to reconstruct EpiP and localize activation origins from BSP, which was later extended to the endocardial surface \cite{journal/TBME/zhou2018}.
Instead of relying solely on BSP, Serinagaoglu \textit{et al.} \cite{journal/TBME/serinagaoglu2006} utilized additional statistical prior information about the EpiP distribution, along with sparse EpiP measurements obtained from multielectrode coronary venous catheters, to enhance EpiP reconstruction within a Bayesian framework.
Erenler \textit{et al.} \cite{journal/MBEC/erenler2019} assumed a time-invariant model when employing the Bayesian MAP for EGM reconstruction, which performed worse than Kalman filter solutions that consider the spatiotemporal variations of EGM.

\subsubsection{Kalman Filter} \label{method:PEA:kalman}

The Kalman filter (KF) is a well-established method for estimating states in dynamic systems and has been used to improve ECGI by incorporating temporal information
\cite{journal/TBME/berrier2004,journal/TBME/ghodrati2006,journal/MBEC/erenler2019}.
It implicitly estimates the electrical activity of the heart based on noisy ECG measurements by modeling the heart as a state-space system \cite{journal/TBME/wang2009b,journal/TBME/liu2010}.
The KF updates estimates of the state $\boldsymbol{u}(\boldsymbol{s}, t)$ using the state transition matrix $\boldsymbol{F}$ and process noise $\boldsymbol{w}_t$ \cite{journal/MBEC/aydin2011},
\begin{equation} 
    \boldsymbol{u}(\boldsymbol{s}, t) = \boldsymbol{F}\boldsymbol{u}(\boldsymbol{s}, t-1) + \boldsymbol{w}_t. 
\end{equation} 
The filter calculates a point estimate and a covariance matrix representing the uncertainty. 
It effectively handles uncertainty by modeling both the system dynamics and measurement noise.
To reduce computational complexity, researchers often focus on temporal relationships among spatially well-correlated regions, such as neighbouring regions or leads belonging to the same activation wavefront.
They may also employ methods like expectation-maximization (EM) and residual-based techniques to estimate noise variances \cite{journal/MBEC/aydin2011}. 
Other formulations, like the Duncan and Horn formulation of the KF used by Berrier \textit{et al.} \cite{journal/TBME/berrier2004}, focus on reconstructing specific EGMs.
Wang \textit{et al.} \cite{journal/TBME/wang2009b}, in an attempt to integrate general prior knowledge with subject-specific data, used a statistical perspective to explicitly account for both model and data errors. 
They represented the cardiac EP system in a state-space form and conducted inverse inference of 3D TMP distributions via the unscented KF.
A similar strategy was employed to define both the electrical and mechanical measurements as the state variables, which can be estimated via the unscented KF \cite{journal/JCP/corrado2015}.
Huang \textit{et al.} \cite{conf/MICCAI/huang2022} proposed a data-driven Kalman filter model, consisting of a state transfer network and a Kalman gain network.
Their method can consider the dynamic activation process of the cardiomyocyters when recovering TMP from BSP.
While the KF is a powerful tool for state estimation in cardiac modeling, it involves large matrices operation and may suffer from the initialization problem.

For EP parameter estimation, to the best of our knowledge, only one study has used a reduced-order unscented KF to estimate cardiac tissue conductivity for personalizing a cardiac EP model \cite{conf/CARS/talbot2015}.
In constrast, KF has been widely employed for the cardiac electromechanical model parameter estimation, such as contractility \cite{journal/JMBBE/xi2011,journal/MedIA/marchesseau2013}.

\begin{figure}[t]\center
 \includegraphics[width=0.5\textwidth]{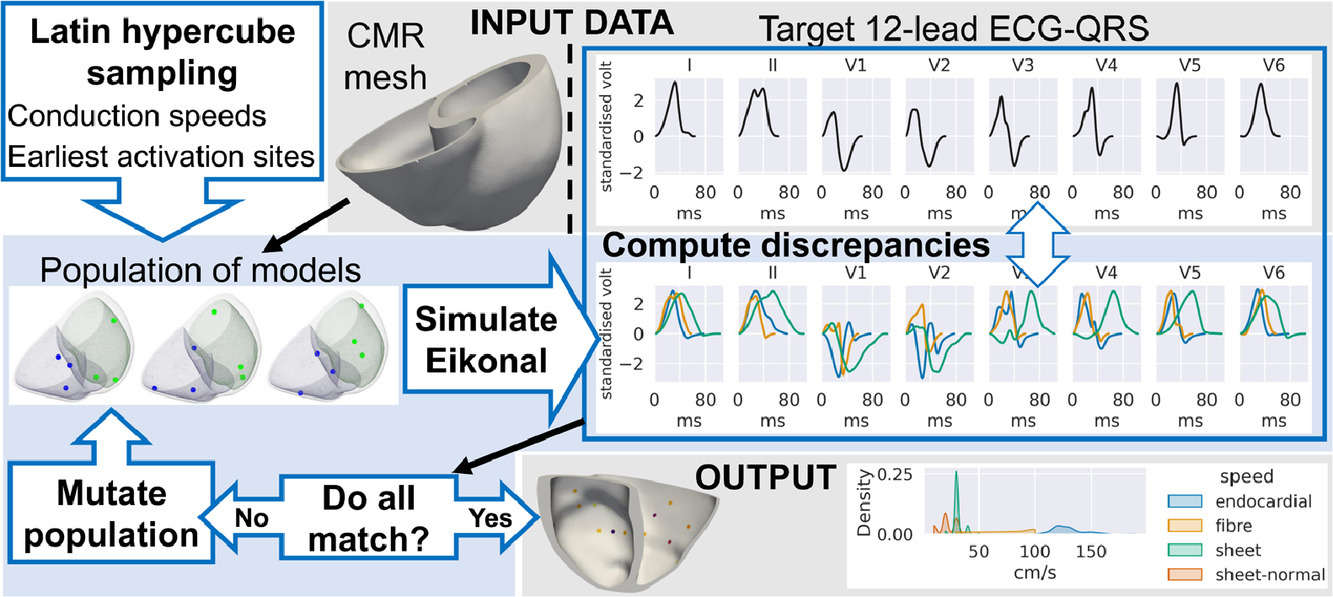}\\[-2ex]
   \caption{Example of the MCMM application on the inverse inference of earliest activation sites (EASs) and CV from ECG. Here, MCMC has been used for the sampling of parameter sets of the population \cite{journal/MedIA/camps2021}. Image adapted from Camps \textit{et al.} \cite{journal/MedIA/camps2021} with permission.
   }
\label{fig:method:MCMC}
\end{figure}

\subsubsection{Markov Chain Monte Carlo Sampling} 

Probabilistic estimation of EP parameters is crucial for model personalization and uncertainty quantification.
Many probabilistic inference methods use Markov Chain Monte Carlo (MCMC) sampling to handle complex models and uncertainties associated with ECG data \cite{journal/FiP/zaman2021}.
For instance, Camps \textit{et al.} \cite{journal/MedIA/camps2021} presented a sequential Monte Carlo approximate Bayesian computation-based model for the inference of ventricular activation properties, as shown in \Leireffig{fig:method:MCMC}.
Rahimi \textit{et al.} \cite{journal/TMI/rahimi2015} applied a hierarchical Bayesian inference to estimate the statistical parameters in the posterior pdf, which can be calculated via MCMM sampling methods.
However, direct MCMC sampling of the pdf of the parameters can be impractical, as it involves repeated evaluations of the posterior pdf involving computationally expensive simulation processes.
To accelerate sampling, various hybrid sampling techniques have been developed using information about the target pdf such as gradient and Hessian matrix \cite{journal/SIAM/martin2012}, which however are challenging to extract when dealing with complex simulation models.
Alternatively, one could develop computationally efficient surrogate models of the expensive simulation process, making it significantly faster to sample the associated pdfs \cite{journal/PBMB/konukoglu2011,journal/NMBE/schiavazzi2016}.
Despite improved efficiency, direct sampling of surrogate-based posterior pdf may yield limited accuracy, given the challenge of creating accurate approximations for complex nonlinear simulation models.
Instead of completely replacing the sampling of the exact posterior pdf, Dhamala \textit{et al.} \cite{journal/MedIA/dhamala2018} proposed a two-stage model to integrate Gaussian process surrogate modeling of the posterior pdf in accelerating sampling.
Also, one could directly develop a surrogate model for the posterior pdf of simulation model parameters, eliminating the necessity for additional MCMC sampling of the original computationally-intensive pdf \cite{journal/FiP/zaman2021}.

\subsubsection{Variational Inference} 

For ECGI, unlike conventional regularization techniques, variational inference (VI) provides a probabilistic framework that can quantify uncertainty in the reconstructed potentials. 
Specifically, it can incorporate prior knowledge about the spatial and temporal dynamics of cardiac activity, improving the robustness of the solution against noisy data.
For example, Ghimire \textit{et al.} \cite{journal/TMI/ghimire2019} introduced a Bayesian framework to jointly infer the posterior distribution of TMP and the errors in the prior EP model from ECG data under the constraint of a 3D EP simulation model.
Xu \textit{et al.} \cite{conf/MICCAI/xu2015} introduced a Bayesian inference framework for robust transmural electrophysiological imaging.
Recently, deep learning-based probabilistic methods have served as an emerging alternative to conventional methods for modeling complex dynamics of cardiac electrical activity \cite{conf/MICCAI/ghimire2018,journal/JCP/yang2021}. 
They can leverage deep neural networks to approximate the posterior distribution of the model parameters or latent variables, providing faster and more accurate approximations. 
Specifically, variational auto-encoder (VAE), as a extension of traditional VI, is quite useful to handle complex, high-dimensional data during the inverse inference. 
For example, Bacoyannis \textit{et al.} \cite{journal/EP/bacoyannis2021} employed a VAE model to reconstruct the activation pattern of the myocardium with various local wall thicknesses as thin walls indicated infarct areas.
Bacoyannis \textit{et al.} \cite{journal/EP/bacoyannis2022} employed conditional VAE to generate activation maps from combined BSP and conductivity maps.

For EP parameter estimation, VI approximates the posterior distribution of EP parameters with a simpler variational distribution by solving an optimization problem.
Similar to ECGI, VAE has also been widely used for EP parameter estimation, such as the inverse inference of ventricular activation properties, i.e., EASs and CVs \cite{conf/STACOM/li2022}.
Graph convolutional VAE can further enhance this estimation by effectively handling non-Euclidean data structures, i.e., cardiac meshes, allowing for more accurate modeling of spatial relationships within the heart \cite{conf/MICCAI/dhamala2019}. 
Specifically, Dhamala \textit{et al.} employed graph convolutional VAE to embed the high-dimensional optimization into low-dimensional latence space for the estimation of tissue excitability \cite{journal/MedIA/dhamala2020}.
Recently, Yang \textit{et al.} \cite{journal/JCP/yang2021} proposed Bayesian PINN to solve linear or nonlinear PDEs with noisy data for both forward and inverse problems.
They employed the Hamiltonian Monte Carlo or the variational inference for the estimation of the posterior distributions.
This approach could be effectively adapted to tackle the ECG inverse problem in the future.



\subsubsection{Surrogate Model for Bayesian Optimization}

To accelerate the inverse inference of parameters, surrogate models, especially statistical emulation with Gaussian processes (GP), have been employed in cardiovascular fluid dynamics \cite{journal/NMBE/melis2017}, the pulmonary circulatory system \cite{conf/CIBB/noe2017}, ventricular mechanics \cite{journal/FiP/di2018}, and LV dynamics \cite{journal/NMBE/melis2017}. 
It involves approximating the computationally intensive mathematical model (the simulator) with a computationally efficient statistical surrogate model (the emulator) \cite{journal/NMBE/melis2017}. 
This could be achieved through a combination of extensive parallelization and GP regression, allowing for optimal utilization of computational resources prior to the patient arrival at the clinic.
GP regression is a non-parametric Bayesian approach used for regression tasks where the model learns from the observed data to make predictions. 
It does not assume a specific parametric form for the underlying function but instead models the function as a distribution over functions, which is determined by the observed data points. 
Therefore, when new data are accessible, the proxy objective function can be minimized at a low computational cost. 
This eliminates the need for additional computationally intensive simulations from the original mathematical model and thus accelerates the inference.
Recently, the GP surrogate model started to be used in the ECG inverse inference by Pezzuto \textit{et al.} \cite{journal/IFAC/pezzuto2022}, where the GP surrogate model with Bayesian optimization has been used for the estimation of EASs. 
Earlier work by Konukoglu \textit{et al.} \cite{journal/PBMB/konukoglu2011} demonstrated Bayesian estimation using polynomial chaos for EASs and conductivity estimation. 
These strategies offer a mid-way solution between pure sampling methods using EP models and machine learning-based solutions as the variational inference.
They retain the flexibility and interpretability of pure sampling methods while replacing the model with a faster emulator, which can be machine learning-based.

\section{Dataset and Evaluation} \label{evaluation}

\begin{figure}[t]\center
 \includegraphics[width=0.3\textwidth]{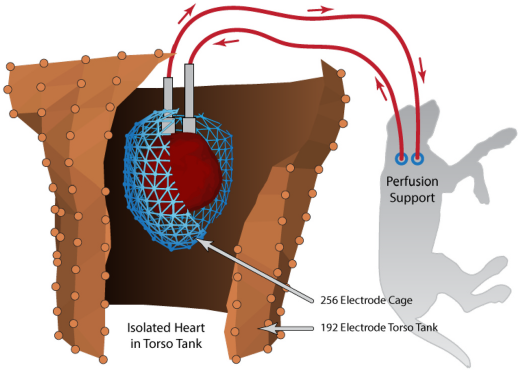}\\[-2ex]
   \caption{Illustration of Utah Torso Tank, with an isolated canine heart. Image adapted from Bergquist \textit{et al.} \cite{journal/CBM/bergquist2021} with permission.
   }
\label{fig:evaluation:utah_tank}
\end{figure}

\begin{figure}[t]\center
 \includegraphics[width=0.48\textwidth]{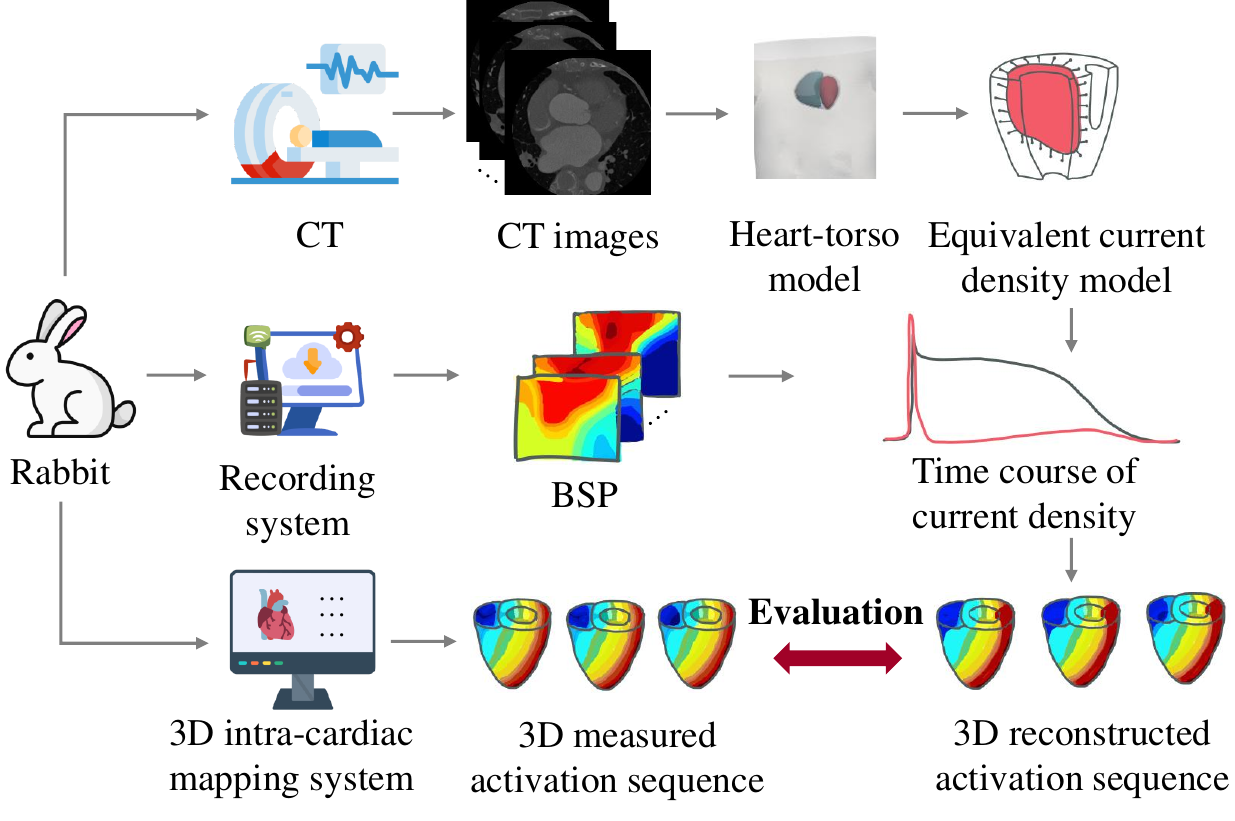}\\[-2ex]
   \caption{Illustration of in-vivo evaluation based on the rabbit undergoing the ventricular pacing from various locations. Illustrations designed referring to Han \textit{et al.} \cite{journal/TMI/han2008}.} 
\label{fig:data:in_vivo_rabbit}
\end{figure}


\subsection{Dataset} 


To evaluate the solution of ECG inverse problem, both anatomical data from cardiac imaging and electrical activity recording data from ECG or BSP are necessary, as shown in \Leireffig{fig:intro:inverseECG}. 
While there are many public cardiac imaging data \cite{journal/MedIA/li2023a,journal/MedIA/zhuang2022,journal/MedIA/zhuang2019,journal/MedIA/li2023b} and ECG recordings \cite{journal/SD/gillette2023,journal/SD/wagner2020,journal/Circulation/goldberger2000}, the evaluation of ECG inverse inference methods requires paired imaging and signal data from the same patient, which is less publicly available.
UK Biobank \cite{journal/EHJ/littlejohns2019} and China Kadoorie Biobank \cite{conf/EMBC/shen2016,journal/IJE/chen2011} provide paired imaging and ECG dataset. 
MyoFit46 includes 500 participants of approximately 75 years+ to undertake high-density surface ECGI and cardiac MRI \cite{journal/BMC/webber2022}. 
Strocchi \textit{et al.} \cite{journal/PO/strocchi2020} provided a publicly available virtual cohort of twenty-four four-chamber hearts built from heart failure patients. 
Experimental Data and Geometric Analysis Repository (EDGAR) is another public dataset for the application and validation of ECGI techniques \cite{journal/JE/aras2015}. 
It includes paired BSPM and endocardial/ epicardial recordings from both animals and torso tank datasets.
Accurate cardiac positioning within the torso is also crucial and often requires torso imaging data for precise localization.
Several studies have attempted to reconstruct the torso directly from standard cardiac imaging for the localization of electrodes, which was challenging due to partial torso shape information \cite{conf/EMBC/smith2022,conf/MICCAI/zacur2017,journal/arxiv/li2024}. 
There are also several validation studies conducted in ex vivo torso tanks \cite{journal/CBM/bergquist2021,journal/FiP/bear2019,journal/HR/bear2018}, in vivo large animal models (rabbit, canine, swine, etc) \cite{journal/CAE/oosterhoff2016,journal/JACC/cluitmans2017,journal/CAE/bear2018,journal/TMI/han2008,journal/HR/han2011}, or in human \cite{journal/HR/ghanem2005,journal/CAE/sapp2012}.
\Leireffig{fig:evaluation:utah_tank} provides the one of the most commonly used ex-vivo torso tanks.
\Leireffig{fig:data:in_vivo_rabbit} presents an in-vivo rabbit model for the evaluation of non-invasive 3D activation sequence reconstruction.


\subsection{Evaluation} 

\begin{table*} [t] \center
    \caption{Summary of selected ECG inverse inference works for ECGI. 
    EGM: electrograms; OVA: origin of ventricular activation; 
    VT: ventricular tachycardia; MI: myocardial infarction; AF: atrial fibrillation; PVC: premature ventricular contraction;
    EpiP/ EndoP: epicardial/ endocardial potential. 
    }
    \label{tb:method:ECGI}
    {\resizebox{0.98\textwidth}{!}{	
    \begin{tabular}{p{0.1cm}|p{3.3cm}|p{2.8cm}p{0.5cm}p{1.5cm}p{2.9cm}p{5cm}}
    \hline
    & Source \& Publish year & Data & Signal & \# Electrodes & Target & Inverse inference method \\
    \hline
    \multirow{40}{*}{\rotatebox{90}{Deterministic estimation}} 
    & Shou \textit{et al.} (2008) \cite{journal/TBME/shou2008}           & in silico          & BSP  & N/A & EpiP       & truncated total least squares \\
    & Wang \textit{et al.} (2009) \cite{journal/TBME/wang2009a}          & in silico          & BSP  & N/A & EpiP       & Tikhonov regularization \\ 
    & Wang \textit{et al.} (2013) \cite{journal/JCP/wang2013}            & in silico          & BSP  & 400 & Myo ischemia, TMP & Tikhonov regularization, total-variation \\ 
    & Cluitmans \textit{et al.} (2018) \cite{journal/MBEC/cluitmans2018} & 3 canines           & BSP  & 192 & ATM, EpiP  & Wavelet regularization \\ 
    & Zhou \textit{et al.} (2018) \cite{journal/JCE/zhou2018}            & 7 VT patients      & ECG; BSP & 120 & OVA   & multi-linear regression; L1-norm regularization \\ 
    & Kalinin \textit{et al.} (2019) \cite{journal/FiP/kalinin2019}      & in silico          & BSP  & N/A & EpiP, EndoP & Tikhonov regularization \\ 
    & Kara \textit{et al.} (2019) \cite{journal/FiP/kara2019}            & in silico          & BSP  & 32/64/128/512/ 1024 & Myo ischemia, EpiP & Tikhonov regularization \\
    & Bear \textit{et al.} (2020) \cite{journal/TBME/bear2020}           & 4 canines/swines     & BSP  & 192/128 & EpiP, ATM  & Tikhonov regularization \\ 
    & Rababah \textit{et al.} (2021) \cite{journal/CBM/rababah2021}      & 6 canines/swines     & BSP  & 192/128 & EpiP, ATM  & Tikhonov regularization  \\    
    & Schuler \textit{et al.} (2021) \cite{journal/TBME/schuler2021}     & in silico          & BSP  & 200 & EpiP, TMP, ATM & Tikhonov regularization \\ 
    \cline{2-7}
    & Messnarz \textit{et al.} (2004) \cite{journal/TBME/messnarz2004}   & in silico          & BSP  & 62  & TMP        & spatiotemporal regularization \\
    & Erem \textit{et al.} (2013) \cite{journal/TMI/erem2013}            & 3 subjects         & BSP  & 120 & EndoP, ATM, pacing sites & temporal spline model \\ 
    & Erem \textit{et al.} (2014) \cite{journal/TMI/erem2014}            & in silico; 1 subject& BSP & 65  & TMP, ATM & temporal spline model \\ 
    & Zhou \textit{et al.} (2016) \cite{journal/TBME/zhou2016}           & 7 AF patients      & ECG  & 208 & atrial EndoP & spatiotemporal frequency based model \\  
    & Yao \textit{et al.} (2017) \cite{journal/JBHI/yao2017}             & 4 MI patients      & BSP  & N/A & MI, EGM    & spatiotemporal regularization \\ 
    & Coll \textit{et al.} (2017) \cite{journal/TBME/coll2017}           & in silico; 1 canine& BSP  & 300, 354 & T-wave alternans, EpiP, EndoP & spatiotemporal spectral and frequency based model \\
    & Yang \textit{et al.} (2018) \cite{journal/TMI/yang2018}            & in silico; 4 swines & BSP  & 208, 128 & ATM        & spatial gradient sparse in frequency domain \\  
    \cline{2-7}
    & Li \textit{et al.} (2001) \cite{journal/TBME/li2001}               & in silico          & BSP  & 200     & OVA        & model optimization based method \\
    & Tilg \textit{et al.} (2002) \cite{journal/TMI/tilg2002}            & 2 patient          & ECG  & 55, 49  & ATM & model optimization based method \\ 
    & Calderero \textit{et al.} (2005) \cite{conf/FIMH/calderero2005}    & 1 canines           & BSP  & 711 & activation wavefront & level set \\
    & Liu \textit{et al.} (2006) \cite{journal/TMI/liu2006}              & in silico          & BSP  & 64/96/128/155/ 200  & ATM  & biophysical model-based method \\ 
    & Han \textit{et al.} (2008) \cite{journal/TMI/han2008}              & 4 rabbits           & BSP  & 53  & ATM        & biophysical model-based method \\ 
    & Chavez \textit{et al.} (2015) \cite{conf/FIMH/chavez2015}          & 3 canines           & BSP  & N/A & Myo ischemia & iterative level set \\
    & Cluitmans \textit{et al.} (2017) \cite{journal/MBEC/cluitmans2017} & 3 canines           & BSP  & 184-216 & EpiP       & physiology-based regularization \\
    \cline{2-7}   
    & Yang \textit{et al.} (2017) \cite{journal/TBME/yang2017}           & in silico; 4 PVC patients & ECG  & 3 & origin of PVC  & data-driven neural network \\ 
    & Karoui \textit{et al.} (2019) \cite{conf/FIMH/karoui2019}          & in silico          & BSP  & N/A & EpiP       & spatial adaptative time delay neural network \\ 
    & Alawad \textit{et al.} (2019) \cite{journal/TMI/alawad2019}        & in silico; 3 patients& ECG & N/A & OVA   & transfer learning \\
    & Gyawali \textit{et al.} (2019) \cite{journal/TBME/gyawali2019}     & 39 VT patients      & ECG  & N/A & OVA        & sequential autoencoder \\
    & Cheng \textit{et al.} (2021) \cite{conf/EMBC/cheng2021}            & in silico          & BSP  & N/A & TMP, MI    & shrinkage-thresholding network \\
    & Mu \textit{et al.} (2021) \cite{conf/MICCAI/mu2021}                & in silico          & BSP  & N/A & TMP, MI, ATM   & shrinkage-thresholding + graph CNN \\ 
    & Camara (2021) \textit{et al.} \cite{journal/FiP/camara2021}        & in silico          & BSP  & 64 & AF driver         & data-driven neural network  \\
    & Karoui \textit{et al.} (2021) \cite{journal/FiP/karoui2021}        & in silico           & BSP  & N/A & atrial ATM & data-driven neural network \\ 
    & Meister \textit{et al.} (2021) \cite{conf/STACOM/meister2021}      & 16 swines            & ECG  & N/A & ATM        & graph convolutional regression \\  
    & Chen \textit{et al.} (2022) \cite{journal/Sensors/chen2022}        & 5 swines            & BSP  & 239 & EpiP, ATM, pacing sites  & data-driven neural network \\   
    & Tenderini \textit{et al.} (2022) \cite{journal/SIAM/tenderini2022} & in silico           & ECG  & 9   & EpiP       & PDE-aware deep learning model \\ 
    & Jiang \textit{et al.} (2024) \cite{journal/TMI/jiang2024}          & in silico; 3 patients& BSP & N/A & EpiP, EndoP & neural state-space modeling \\
    & Lian \textit{et al.} (2024) \cite{journal/TBME/lian2024}           & in silico; 4 patients&ECG & 9   & MI         & frequency-enhanced network \\
    \cline{2-7}
    & Jiang \textit{et al.} (2022) \cite{journal/TMI/jiang2022}          & 1 canine           & BSP  & 192 & EpiP       & physics embedded graph CNN  \\
    & Xie \textit{et al.} (2023) \cite{journal/TASE/xie2023}             & in silico          & BSP  & N/A & EpiP       & physics-constrained network \\
    \hline \hline
    \multirow{15}{*}{\rotatebox{90}{Probabilistic estimation}} 
    & Serinagaoglu \textit{et al.} (2006) \cite{journal/TBME/serinagaoglu2006} & 1 canine & BSP  & 771 & EpiP & Bayesian MAP \\ 
    & Wang \textit{et al.} (2010) \cite{journal/TBME/wang2010}           & in silico          & BSP  & N/A & TMP, MI & Bayesian MAP \\
    & Zhou \textit{et al.} (2018) \cite{journal/TBME/zhou2018}           & in silico; 3 patients   & BSP  & 120 & EndoP, pacing sites & sparse Bayesian MAP \\    
    & Zhou \textit{et al.} (2018) \cite{journal/TBME/zhou2018b}          & in silico; 3 VT patients& BSP  & 120 & EpiP    & spatio-temporal Bayesian MAP \\    
    \cline{2-7}
    & Wang \textit{et al.} (2009) \cite{journal/TBME/wang2009b}          & in silico; 1 MI patient & BSP  & 700/330, 123 & TMP        & physiological-constrained Kalman model \\ 
    & Huang \textit{et al.} (2022) \cite{conf/MICCAI/huang2022}          & in silico          & BSP  & N/A & TMP        & data-driven Kalman filtering network \\
    \cline{2-7}
    & Wang \textit{et al.} (2010) \cite{conf/CiC/wang2010}               & 1 canine           & BSP  & N/A & EpiP       & variational-form-based regularizers \\ 
    & Wang \textit{et al.} (2011) \cite{journal/TBME/wang2011}           & 1 canine           & BSP  & N/A & EpiP       & variational regularizers \\   
    & Xu \textit{et al.} (2014) \cite{conf/MICCAI/xu2014}                & in silico; 6 patients& BSP& N/A & MI         & variational Bayesian \\  
    & Figuera \textit{et al.} (2016) \cite{journal/FiP/figuera2016}      & in silico          & BSP  & N/A & atrial EpiP & Bayesian MAP, etc \\
    & Ghimire \textit{et al.} (2019) \cite{journal/TMI/ghimire2019}      & in silico; 2 patients & ECG & N/A & TMP      & Bayesian model \\
    & Ozkoc \textit{et al.} (2021) \cite{conf/ICM/ozkoc2021}             & 1 canine           & BSP  & 192 & EpiP       & Bayesian MAP \\
    & Li \textit{et al.} (2024) \cite{journal/TMI/li2024}                & in silico          & ECG  & 10 & MI         & multi-modal VAE  \\ 
    \hline
\end{tabular}} 
}
\end{table*}

\begin{table*} [t] \center
    \caption{Summary of the selected ECG inverse inference works for EP parameter estimation. HF: heart failure; APD: action potential duration; MS: Mitchell-Schaeffer; AP: Aliev-Panfilov.
     }
    \label{tb:method:EP parameter}
    {\resizebox{0.98\textwidth}{!}{	
    \begin{tabular}{p{0.1cm}|p{3.4cm}|p{2.7cm}p{0.5cm}p{3.5cm}p{2cm}p{4.2cm}}
    \hline
    & Source \& Publish year & Data & Signal & Target & EP model & Inverse inference method \\
    \hline
    \multirow{18}{*}{\rotatebox{90}{Deterministic estimation}} 
    & Chinchapatnam \textit{et al.} (2008) \cite{journal/TMI/chinchapatnam2008} & in silico; 1 patient & N/A  & local CV; conductivity & Eikonal & adaptive zonal iterative model \\ 
    & Sermesant \textit{et al.} (2012) \cite{journal/MedIA/sermesant2012}& 2 patients          & ECG  & local CV, EAS    & Eikonal       & model-based regularization \\ 
    & Potse \textit{et al.} (2014) \cite{journal/EP/potse2014}           & 2 HF patients       & ECG  & EAS, conductivity, etc & Ten Tusscher and Panfilov       & model-based regularization \\ 
    & Kahlmann \textit{et al.} (2017) \cite{journal/CDBE/kahlmann2017}   & in silico           & ECG  & Purkinje network& Eikonal        & parallel simplex optimization \\
    & Sanchez \textit{et al.} (2018) \cite{journal/MBEC/sanchez2018}     & 6 patients          & ECG  & EAS, conductivity, etc & Ten Tusscher and Panfilov & iterative method \\
    & Lee \textit{et al.} (2019) \cite{journal/MedIA/lee2019}            & in silico           & ECG  & global CV     & reaction-Eikonal & rule based model \\
    & Pezzuto \textit{et al.} (2021) \cite{journal/EP/pezzuto2021}       & 11 HF patients      & ECG  & global CV, EAS& Eikonal          & propagation model \\
    & Grandits \textit{et al.} (2021) \cite{journal/JNMBE/grandits2021}  & in silico; 1 patient& ECG  & EAS        & Eikonal           & topological gradient  \\ 
    & Costa \textit{et al.} (2022) \cite{journal/CBM/costa2022}          & 6 pigs              & ECG  & global CV, APD, conductivity & reaction-Eikonal & total activation time fit \\
    & Grandits \textit{et al.} (2023) \cite{journal/TBME/grandits2023}   & 1 patient           & ECG  & local CV, EAS, ATM & Eikonal & Geodesic backpropagation  \\
    \cline{2-7}
    & Neumann \textit{et al.} (2016) \cite{journal/MedIA/neumann2016}    & in silico           & ECG  & global CV, Purkinje network & Eikonal & reinforcement learning \\
    & Giffard \textit{et al.} (2018) \cite{journal/TBME/giffard2018}     & 20 patients         & BSP  & local CV, pacing site, ATM & MS   & transfer learning \\
    \cline{2-7}
    & Herrero \textit{et al.} (2022) \cite{journal/FCM/herrero2022}      & in silico; in vitro & ECG 
    & APD, excitability and diffusion coefficients & AP    & PINN \\ 
    & Ye \textit{et al.} (2023) \cite{conf/MICCAI/ye2023}                & in silico           & ECG  & tissue excitability & AP & PINN \\
    \hline \hline
    \multirow{11}{*}{\rotatebox{90}{Probabilistic estimation}} 
    & Zettinig \textit{et al.} (2014) \cite{journal/MedIA/zettinig2014}  & in silico  & ECG  & electrical diffusivity & MS & statistical polynomial regression \\
    & Gillette \textit{et al.} (2021) \cite{journal/MedIA/gillette2021}  & 1 patient           & ECG  & global CV, Purkinje network, APD, conductivity & reaction-Eikonal & Saltelli sampling \\
    & Camps \textit{et al.} (2021) \cite{journal/MedIA/camps2021}        & in silico           & ECG  & global CV, EAS& Eikonal          & Monte Carlo Bayesian inf \\
    & Camps \textit{et al.} (2024) \cite{journal/MedIA/camps2024}        & in silico           & ECG  & global CV, Purkinje network & Eikonal & Monte Carlo Bayesian inf \\
    \cline{2-7} 
    & Zaman \textit{et al.} (2021) \cite{journal/FiP/zaman2021}          & in silico; 3 VT patients & ECG & tissue excitability & AP  & Bayesian active learning \\
    & Dhamala \textit{et al.} (2020) \cite{journal/MedIA/dhamala2020}    & in silico; 3 VT patients & ECG  & tissue excitability & AP  & generative VAE \\ 
    & Li \textit{et al.} (2022) \cite{conf/STACOM/li2022}                & 1 patient           & ECG  & global CV, EAS& Eikonal          & conditional VAE \\
    \cline{2-7}  
    & Konukoglu \textit{et al.} (2011) \cite{journal/PBMB/konukoglu2011} & in silico; 1 patient& BSP  & EAS, conductivity& Eikonal       & polynomial chaos based Bayesian inf \\
    & Dhamala \textit{et al.} (2017) \cite{journal/TMI/dhamala2017}      & in silico; 2 VT patients & ECG  & tissue excitability & AP  & spatially adaptive surrogate model \\ 
    & Pezzuto \textit{et al.} (2022) \cite{journal/IFAC/pezzuto2022}     & in silico           & ECG  & EAS & Eikonal & GP based Bayesian optimization \\ 
    \hline
\end{tabular}} 
}
\end{table*}

\begin{figure}[t]\center
 \includegraphics[width=0.5\textwidth]{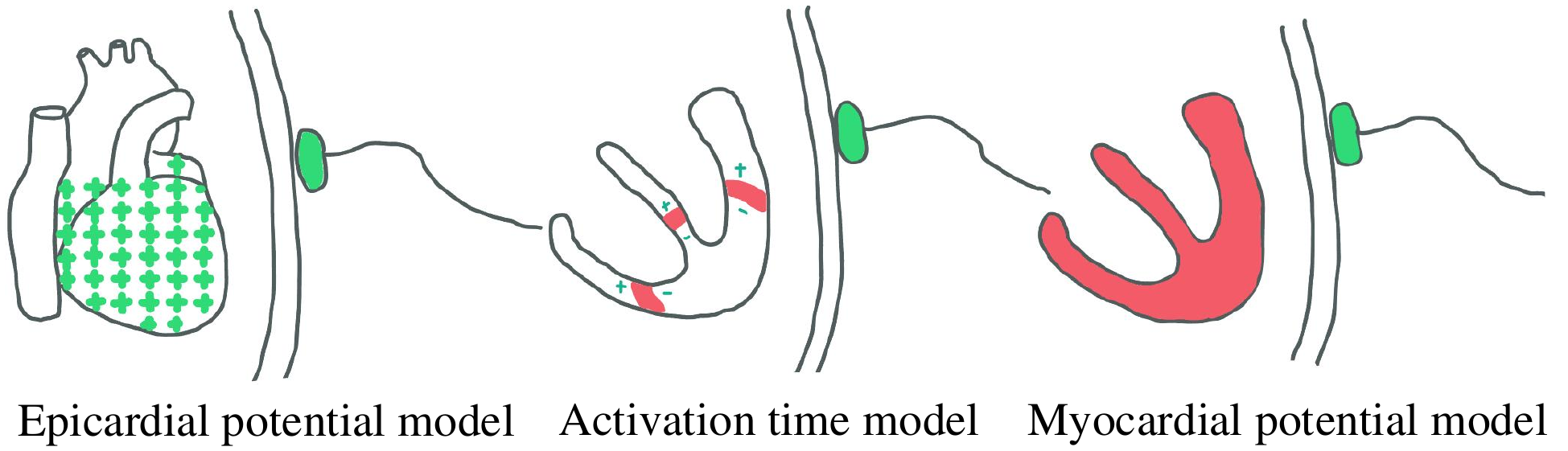}\\[-2ex]
   \caption{Illustration of different cardiac source models in ECGI. 
   }
\label{fig:eval:source models}
\end{figure}

The validation of the inverse problem involves a diverse set of targeted cardiac sources or EP parameters, input signal, forward EP model formulations, and inverse inference methods, as presented in \Leireftb{tb:method:ECGI} and \Leireftb{tb:method:EP parameter}. 
Note that the tables only provide a selection of representative studies and are not an exhaustive list of all available studies.
One can see that BSP has been widely used for reconstructing cardiac electrical activity, while ECG is commonly used for EP parameter estimation.
Several studies have shown that the number of electrodes greatly affects the accuracy of cardiac source reconstruction \cite{journal/TMI/liu2006,journal/FiP/kara2019}. 
Instead of using fixed number of electrodes, Cluitmans \textit{et al.} \cite{journal/MBEC/cluitmans2017} proposed determining the number based on the torso size of the individual.
Alternatively, Ondrusova \textit{et al.} \cite{journal/FiP/ondrusova2023} proposed a two-step single dipole based inverse inference model, initially using all electrodes and then refining with a subset selected by significance via a greedy algorithm.
For EP parameter estimation, the AP and (reaction-)Eikonal models are frequently used due to their balance between computational efficiency and accurate representation of cardiac activation dynamics. 
The Mitchell-Schaeffer (MS) model, while less commonly used, offers simplicity and is suitable for specific scenarios requiring less complexity \cite{journal/MedIA/zettinig2014,journal/TBME/giffard2018}.
Most studies employed computer models for in silico evaluation \cite{journal/TBME/schuler2021,journal/FiP/kalinin2019} and/ or performed evaluation on the basis of torso-tank experiments with isolated animal hearts, such as rabbit \cite{journal/AJPHCP/zhang2005,journal/TMI/han2008}, canine \cite{conf/FIMH/calderero2005,journal/MBEC/cluitmans2018,journal/TMI/jiang2022}, and swine \cite{conf/STACOM/meister2021,journal/TBME/throne2002,journal/TMI/yang2018,journal/Sensors/chen2022}.
Also, there exist several in-situ animal studies, where the potential of the heart and body surfaces were simultaneously recorded in animals, such as Auckland pig or Utah/ Maastricht dog data \cite{journal/CAE/bear2018,journal/TBME/serinagaoglu2006,journal/JACC/cluitmans2017}.
There are also few studies evaluated using in-vivo human subjects, which however usually employed non-simultaneous invasive recordings, such as EGMs \cite{journal/EP/pezzuto2021} or bipolar voltages maps \cite{conf/MICCAI/xu2014,journal/TMI/ghimire2019}.
The comparison with the published invasive measurements in animals or humans has also been utilized as an alternative evaluation \cite{journal/Nature_medicine/ramanathan2004}.
As for inference target, for ECGI, there are three major cardiac source models, i.e., surface potential models (EpiP and EndoP), activation models, and transmembrane voltage models, as shown in \Leireffig{fig:eval:source models}.
Additionally, some studies have focused on reconstructing regions of electrical conduction abnormalities \cite{journal/JBHI/yao2017,conf/MICCAI/mu2021,journal/TMI/li2024,journal/TBME/yang2017} and the origins of ventricular activation \cite{journal/JCE/zhou2018,journal/TMI/alawad2019,journal/TBME/gyawali2019}.
In general, reconstruction efforts have primarily focused on the ventricles, with relatively few studies addressing atrial imaging, as specifically highlighted in \Leireftb{tb:method:ECGI}.
This might be due to the atrial thinner walls, irregular anatomy, and more complex electrical activity \cite{journal/FiP/salinet2021}.
For EP parameter estimation, EAS (or Purkinje network), CV, conductivity, and tissue excitability have been mainly considered.


\subsubsection{Surface Potential Model}

As summarized in \Leireftb{tb:method:ECGI}, surface potentials, specifically EpiP and EndoP, are the most common targets for cardiac source reconstruction. 
The prevalence can be attributed to its practicality and clinical significance, with EpiP being particularly favored as it is more accessible for measurement than EndoP.
Furthermore, surface potentials usually require less computational power compared to full-volume models. 
They are generally faster to functionalize since they deal with surface data, and often rely on established methods and datasets.
Nonetheless, the reconstruction of surface potentials can suffer from non-uniqueness, particularly if the input data is not sufficiently detailed.

\subsubsection{Activation Time Model}

ATM is commonly used to describe the propagation of electrical activity across the heart. 
It is a simplified model focused on the activation phase and assumes a smooth, single wavefront, which limits its applicability to more complex cardiac events, such as reentry and arrhythmias.
The functionalization of ATM is relatively quick as it focuses on timing rather than detailed voltage dynamics, simplifying the modeling process.
Most studies have reconstructed the ATM on the atrial and/or ventricular surfaces from BSP or ECG \cite{journal/TMI/tilg2002,journal/FiP/karoui2021,journal/MBEC/cluitmans2018,journal/CBM/rababah2021}. 
Instead, some studies have reconstructed the 3D cardiac activation sequence throughout the myocardial volume, evaluated both in silico \cite{journal/TMI/liu2006,journal/TBME/schuler2021} and in vivo, including rabbit and canine models \cite{journal/TMI/han2008,journal/AJPHC/han2012,journal/HR/han2011}.
For example, Schuler \textit{et al.} \cite{journal/TBME/schuler2021} conducted an in-silico study to identify best conﬁgurations for reducing line-of-block artifacts in ATM reconstructed from EpiP or TMP.
Han \textit{et al.} \cite{journal/TMI/han2008} performed an in vivo validation study by comparing reconstructed activation imaging with 3D intracardiac mapping in a group of four healthy rabbits undergoing ventricular pacing from different locations, as presented in \Leireffig{fig:data:in_vivo_rabbit}.
The reconstruction of ATM can be more unique in terms of identifying activation sequences but might lack the details needed to reconstruct the full electrical activity accurately.

\subsubsection{Transmembrane Potential Model}

Using TMP as a cardiac source model provides an accurate approximation of the cardiac electrical activity, revealing critical arrhythmic details and intramural arrhythmogenic substrates \cite{journal/FiP/cluitmans2018}. 
However, this approach is computationally demanding due to the need for complex 3D modeling and extensive data (volumetric mesh, fibre orientation, myocardial conductivities, etc), leading to longer setup and functionalization times. 
Despite these challenges, TMP reconstructions are valuable for identifying depolarized and repolarized regions and detecting ischemic, fibrotic, or border-zone areas. 
Since TMPs cannot be directly measured clinically, validation relies on extracellular signals generated by TMP distributions. 
Surface potentials and ATM can be derived from TMP distributions and then measured in experiments.

\subsubsection{EP Parameter}

As summarized in \Leireftb{tb:method:EP parameter}, the parameters to be personalized are directly tied to the forward EP models. 
For instance, the Eikonal model typically involves the personalization of parameters such as the EAS, CV, and tissue conductivity \cite{journal/TMI/chinchapatnam2008,journal/MedIA/camps2021,journal/MedIA/camps2024,journal/MedIA/gillette2021}. 
For example, Camps \textit{et al.} \cite{journal/MedIA/camps2021} used dynamic time warping (DTW) and Pearson's correlation coefficients to align simulated ECG with clinical acquired ECG, enabling the iterative updating of EASs and CVs, as shown in \Leireffig{fig:method:MCMC}.
In contrast, the primary parameter optimized in AP models is usually tissue excitability \cite{journal/TMI/dhamala2017,journal/MedIA/dhamala2020,journal/FiP/zaman2021,conf/MICCAI/ye2023}. 
However, these are not the only parameters that may require optimization. 
Depending on the specific application and model, additional parameters such as ionic channel properties, diffusion coefficients, or APD might also need to be personalized to accurately capture patient-specific cardiac behavior \cite{journal/MedIA/gillette2021,journal/FCM/herrero2022}. 
The choice of parameters to optimize is ultimately guided by the intended clinical application and the complexity of the model, balancing between computational feasibility and the need for precision in capturing the physiological phenomena of interest.

\subsection{Result} 


Validating the results of ECG inverse inference is highly challenging due to the lack of a unified or standardized pipeline for ECGI or EP parameter estimation. 
The most commonly used evaluation metric for potential and activation models is the correlation coefficient (CC) \cite{journal/CAE/bear2018, journal/sensors/wang2023}. 
For assessing the performance of pacing site, PVC origin, or cardiac excitation origin localization, the Euclidean or geodesic distance between the exact and reconstructed locations is typically employed \cite{journal/FiP/karoui2021, journal/Sensors/chen2022, journal/AIM/pilia2023, journal/FiP/dogrusoz2023}.
As mentioned earlier, validation methods, source models, and the complexity of the forward model (e.g., homogeneous vs. inhomogeneous torso models) can vary significantly, leading to significant differences in results reported across studies.
Generally, most previous studies have reported a median CC of approximately 0.7 for EpiP reconstruction, and this value can reach 0.8 to 0.9 for ATM and TMP reconstructions \cite{journal/Sensors/chen2022, conf/ISBI/xie2021}. 
Pacing site localization errors can range from 2.0 to 20.0 mm, with a median distance of around 10 mm \cite{journal/Sensors/chen2022, conf/MICCAI/mu2021,journal/FiP/dogrusoz2023}. 
A recent study compared the pacing site localization performance of dipole and potential-based source models using the public EDGAR dataset, reporting a median distance of 25.2 mm and 13.9 mm, respectively \cite{journal/FiP/dogrusoz2023}.
For PVC origin localization, they reported errors of 30.2-33.0 mm and 28.9-39.2 mm for the dipole and potential-based model, respectively. 
The latest results for cardiac excitation origin localization reported an error of about 23.9 mm on clinical data after fine-tuning, and 44.0 mm without fine-tuning \cite{journal/TMI/jiang2024}.
This finding highlights the challenges of generalization when training ECGI reconstruction networks supervised by simulation data. 
Similar conclusions were drawn by Pilia \textit{et al.} \cite{journal/AIM/pilia2023}, who reported an error of about 1.5 mm on simulated data, which increased to 32.6 mm on unseen clinical data.
ECGI has also been utilized to detect infarcted regions, achieving a average Dice score of 0.4 on simulated data \cite{journal/TMI/ghimire2019,journal/TMI/li2024}. 
EP parameter estimation involves determining intrinsic cardiac properties that are less directly observable, making its evaluation more complex. 
Furthermore, it requires sophisticated models to capture the underlying biophysical processes.
Given these challenges, its validation often requires indirect methods, such as comparing simulation results against observed clinical measurement \cite{journal/MedIA/camps2021}.
Some studies use a combination of in-silico modeling and in vivo data, employing metrics such as root mean squared error (RMSE) and Dice coefficient for evaluation \cite{journal/FiP/zaman2021}. 
These findings highlight the significant challenges in ECG inverse inference evaluation, emphasizing the need for standardized validation methods and improved model robustness.

\section{Clinical Application} \label{clinical_application}

While ECGI and EP parameter estimation focus on different aspects, i.e., detailed electrical imaging and personalized physiological modeling, respectively, they can be effectively integrated within the broader context of CDTs. 
CDTs leverage personalized data to create a comprehensive virtual heart model, encompassing detailed anatomical, electrophysiological, and dynamic information.  
It can not only optimize diagnostic and treatment planning, as with conventional ECGI, but also support long-term risk assessment and personalized disease management.
Next, we will introduce the clinical applications of ECGI and the additional advantages brought by CDTs driven by personalized EP parameter estimation.
For a more comprehensive overview of clinical applications, we refer readers to existing reviews on ECGI \cite{journal/FiP/cluitmans2018, journal/FiP/salinet2021} and CDTs \cite{journal/NRC/niederer2019,journal/JAHA/sel2024}.

\subsection{Preoperative Planning for Cardiac Surgery}

The ability of ECGI to non-invasively characterize cardiac activation patterns is crucial for preoperative planning, such as catheter ablation.
For example, ECGI has been effectively used to guide ablation in patients with persistent AF, achieving success rates of up to 85\% of patients being free from AF 12 months post-procedure \cite{journal/Cir/haissaguerre2014}.
However, ablation procedures are not totally free of risks, require significant resources, and are not suitable for all AF patients \cite{journal/CAE/cappato2010}.
Supporting the potential role of ECGI in patient selection, Meo \textit{et al.} \cite{journal/FiP/meo2018} demonstrated that ECGI could help identify patients likely to benefit from ablation. 
Similarly, Rodrigo \textit{et al.} \cite{journal/CAE/rodrigo2020} showed that ECGI could stratify patients by projecting the complexity of electrical patterns onto the atrial surface, recommending ablation for those with less complex substrates.
Despite these advances, clinical adoption of ECGI for AF ablation remains limited due to the incomplete understanding of AF mechanisms and ablation effects. 
In contrast, ECGI has gained more clinical acceptance for VT ablation \cite{journal/NEJM/cuculich2017}.
Netherless, in either the atrial or ventricle EP procedure setting, the cardiac imaging used in ECGI need to be better integrated with EP to optimize its usage. 

\begin{figure}[t]\center
 \includegraphics[width=0.5\textwidth]{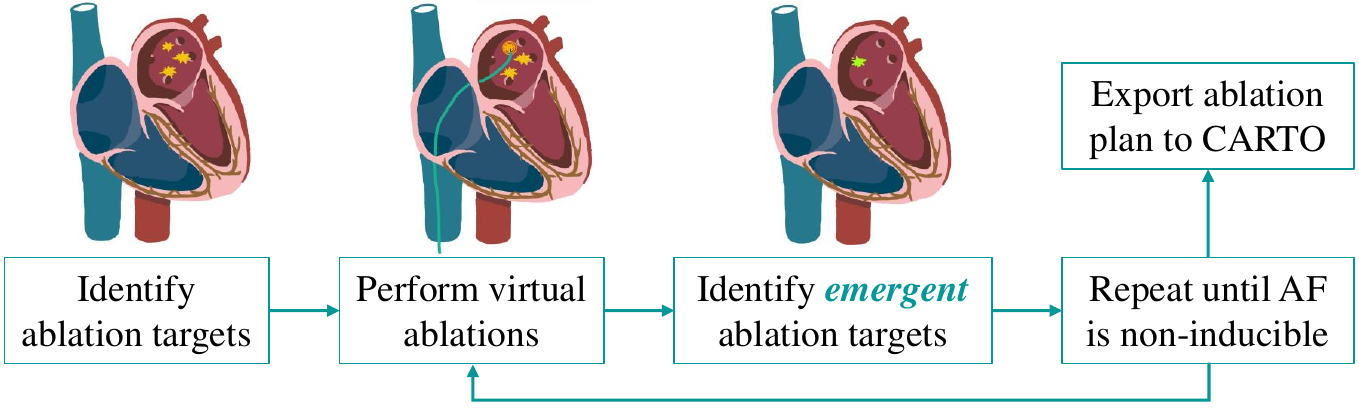}\\[-2ex]
   \caption{Virtual heart model guided preoperative planning for the personzalized catheter ablation of AF patients. Here, CARTO is the clinical electroanatomic navigation system. Illustrations designed referring to Boyle \textit{et al.} \cite{journal/Nature_BME/boyle2019}.
   }
\label{fig:application:AF}
\end{figure}

CDTs expand upon the capabilities of ECGI by providing a more comprehensive simulation environment for patient-specific planning. 
It enable the simulation of various surgical interventions, allowing surgeons to predict outcomes, optimize strategies, and assess risks in a fully controlled, risk-free setting.
For instance, it has been applied to identify optimal infarct-related ablation targets for VT \cite{journal/Nature_BME/prakosa2018,journal/Europace/bhagirath2023}.
This can be accomplished by performing virtual multi-site delivery of electrical stimuli (pacing) from various bi-ventricular locations. 
Each location attempts to induce VT from a site positioned differently relative to the infarct area.
The optimal ablation lesions can then be determined in each personalized virtual heart, rendering it non-inducible for VT from any pacing location.
Therefore, it can terminate not only clinically manifested or induced VTs during the procedure but also all potential VTs arising from the post-infarction substrate. 
This includes VTs that may emerge after the initial ablation, potentially eliminating the need for repeated procedures and providing long-term freedom from VT for patients.
Similarly, this technology allows for the preoperative identification of optimal ablation targets for AF \cite{journal/Nature_BME/boyle2019}, as shown in \Leireffig{fig:application:AF}. 
Additionally, CDTs hold the potential for the virtual implantation of cardiac devices, such as pacemakers, defibrillators, or transcatheter valves, to evaluate their placement and function \cite{journal/NRC/niederer2019}. 
Complex surgical procedures, including coronary artery bypass grafting and valve repair or replacement, can also be simulated to optimize strategies, minimize risks, and improve patient outcomes \cite{journal/BMM/ballarin2017,journal/JACC/de2016}.

\begin{figure*}[t]\center
 \includegraphics[width=0.76\textwidth]{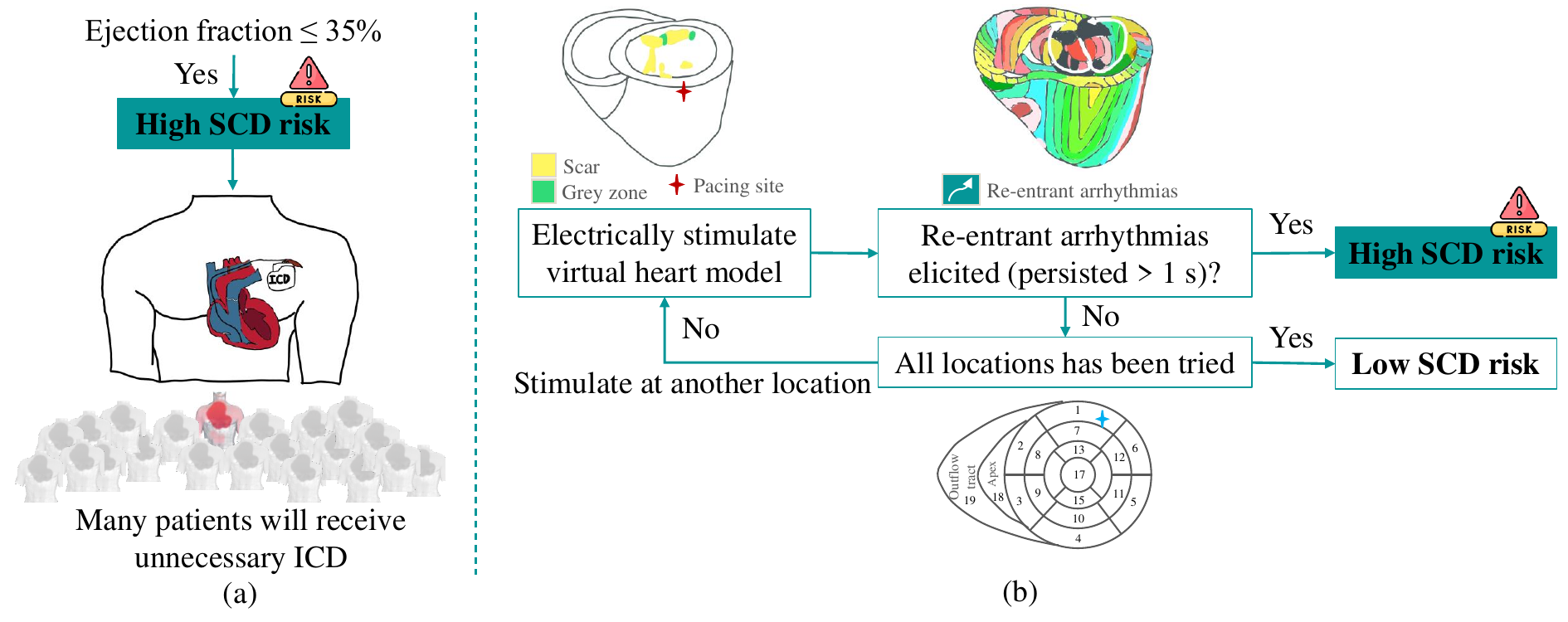}\\[-2ex]
   \caption{Comparison of conventional clinical metric and virtual heart model based identification of patients at high sudden cardiac death (SCD) risk.
   (a) a universal metric in clinical cardiology practice,
   (b) a personalized virtual-heart arrhythmia risk predictor.
   ICD: implantable cardioverter deﬁbrillator.
   Illustrations designed referring to Arevalo \textit{et al.} \cite{journal/Nature_C/arevalo2016}.
   }
\label{fig:application:ICD}
\end{figure*}

\subsection{Risk Stratification and Long-Term Monitoring}

By imaging and quantifying substrates associated with increased risk of adverse cardiac events, ECGI has been used for risk stratification \cite{journal/JACC/cuculich2011}. 
Early studies have identified key substrates for arrhythmias, such as slow discontinuous conduction pathways \cite{journal/Cir/zhang2015}, abnormal repolarization \cite{journal/JAHA/bear2021}, and steep repolarization gradients \cite{journal/Cir/vijayakumar2014}.

CDT model provides a more comprehensive, personalized approach for risk stratification.
It integrates the personalized geometry of the patient, structural remodeling resulting from diseases and electrical functions from sub-cellular to organ levels.
For instance, a virtual-heart arrhythmia risk predictor (VARP) has been created for sudden cardiac death (SCD) based on a virtual heart model \cite{journal/Nature_C/arevalo2016}, as shown in \Leireffig{fig:application:ICD}.
Within the model, a virtual multi-site delivery of electrical stimuli originating from ventricular locations at varying distances to the remodelled tissue can be performed. 
It has been demonstrated to significantly outperform clinical metrics in predicting future arrhythmic events.
The virtual heart model offers significant advantages over ECGI by allowing for non-invasive, personalized predictions of SCD risk, potentially preventing unnecessary implantable cardioverter-defibrillator (ICD) placements in post myocardial infarction patients.
Moreover, it enables continuous monitoring of patients with known cardiovascular conditions, allowing for early detection of deterioration and timely adjustments to treatment plans \cite{conf/CCSSE/rivera2019,conf/MeMeA/martinez2019}. 
By integrating data from multiple sources, such as fitness trackers, wearables, and companion apps, CDTs enhance the utility of ECG inverse inference, enabling more proactive and individualized cardiovascular care.

\subsection{Personalized Disease Insight and Diagnosis}

Numerous studies have demonstrated the potential of ECGI for understanding disease mechanisms and identifying clinically relevant parameters \cite{journal/HR/intini2005,journal/IJC/rudy2017,journal/JCTR/gray2018}.
One of the earliest and most explored clinical applications of ECGI is in the precise identification of arrhythmia mechanisms to guide ablation therapy \cite{journal/FiP/cluitmans2018}. 
Also, ECGI has been used to delineate the distinct activation patterns during arrhythmia (e.g., focal vs. re-entrant) \cite{journal/Cir/cuculich2010,journal/JCE/roten2012}, and to identify the primary drivers of these patterns, which often correlate with successful ablation targets \cite{journal/JCE/pedron2016, journal/TBME/zhou2016}.
Beyond atrial applications, it has also been effective in detecting and characterizing arrhythmogenic substrates that contribute to the onset and maintenance of VT \cite{journal/IJC/rudy2017, journal/PCE/zhang2016} and other ventricular arrhythmias \cite{journal/STM/wang2011}
It has been shown to improve patient selection for CRT by identifying greater ventricular electrical dyssynchrony among responders compared to standard 12-lead ECG methods \cite{journal/JE/varma2007,journal/JACC/ploux2013}. 
The growing commercial availability of ECGI systems, such as CardioInsight (Medtronic, USA), Corify Care, Vivo (Catheter Precision), and EP Solutions SA, further highlights its potential for clinical use \cite{journal/EHJ/corral2020,journal/MBEC/onak2019}.

At the same time, CDTs are transforming the landscape of in silico clinical trials, particularly for evaluating treatment effects in patients with rare diseases, where traditional trials have significant limitations \cite{journal/EHJ/corral2020}.
By enabling highly personalized simulations that capture complex cardiac dynamics and patient-specific characteristics, CDTs enhance the generalizability and reliability of trial results while reducing time and costs \cite{journal/MedIA/camps2024, journal/DDT/moingeon2023}. 
This technique accelerates drug development and provides deeper insights into drug effects, interactions, and mechanisms.
While these studies have significantly advanced our understanding of cardiac disease mechanisms and identified potential therapeutic targets, further prospective studies are needed to validate these insights in clinical settings and explore their potential for guiding diagnosis and patient selection.

\section{Discussion and Future Perspectives} \label{discussion}

Here, we provide a structured discussion on the current challenges in ECG inverse inference and propose potential solutions to address these challenges. 
By evaluating uncertainties, exploring computationally efficient models, and integrating multimodal data, we aim to provide a comprehensive roadmap for future research in this field.

\subsection{Quantifying and Mitigating Uncertainties in ECG Inverse Inference} \label{discussion:UQ}


Despite significant advances in techniques used for ECG inverse inference, uncertainty remains inadequately quantified across many aspects of the pipeline.
Uncertainty quantification (UQ) aims to quantify the range of possible outcomes and their associated probabilities due to input variability \cite{journal/JP/mirams2016}. 
Recently, UQ has been widely employed in cardiac EP modeling, primarily for forward problems, to examine how variations in geometry and EP parameters affect the simulated ECG morphology \cite{conf/FIMH/multerer2021,conf/FIMH/li2023,journal/arxiv/zappon2024}.
Instead, several studies have focused on how input variations inversely affect the reconstructed cardiac sources or estimated EP parameters, namely inverse UQ. 
These is generally achieved via polynomial chaos expansion, which allows for a compact and efficient representation of the uncertainty in model outputs \cite{conf/FIMH/tate2021,conf/CiC/jiang2022,journal/PM/bergquist2023}.
Furthermore, probabilistic approaches, which are often used for inverse UQ, naturally trace uncertainty from observed data back to the model estimation. 
ECG inverse problems are typically ill-posed, often lacking unique solutions, especially in complex scenarios such as activation reconstruction or model personalization incorporating CV and Purkinje networks. 
While uniqueness can sometimes be recovered in standard ECG imaging (the linear case), it is less likely in these more advanced models.
However, clinicians typically require a single, reliable prediction for decision-making in patient care. 
Therefore, it is crucial to quantify and mitigate various uncertainties to enhance the robustness and clinical applicability of ECG inverse solutions.
As illustrated in the ECG inverse inference pipeline presented in \Leireffig{fig:intro:inverseECG}, the primary sources of uncertainty include measurement errors, anatomical modeling inaccuracies, and assumptions in computational modeling. 

ECG or BSP data typically exhibit sparsity with measured values associated with a higher degree of uncertainty.
For instance, Njeru \textit{et al.} \cite{journal/CMBBE/njeru2022} quantified the uncertainty on the reconstructed EpiP due to the ECG measurement.
Moreover, there is ambiguity in matching the BSPM measurement points with the computational mesh due to the non-synchronized BSPM and imaging acquisition \cite{journal/PBMB/konukoglu2011}.
The cardiac position is also susceptible to various sources of movement artifacts, including respiration and cardiac contraction, introducing geometric errors into the ECGI solutions \cite{journal/TBME/jiang2008}.
The geometric error in the anatomical modeling (forward model) should also be considered, as it will in turn cause estimation errors in the ECG inverse inference \cite{journal/TBME/greensite2003,journal/MBEC/aydin2011,journal/FiP/coll2018}.
The anatomical modeling is normally started with image segmentation, so several studies directly performed UQ to investigate the effects of inaccurate segmentation on the reconstructed cardiac source \cite{conf/FIMH/tate2021,conf/CiC/gassa2022}. 
Incorporating a prior model can mitigate uncertainties, and a combination of prior models may be preferred for estimating complex source structures \cite{journal/TMI/rahimi2015}.
Also, one can disentangle the inter-subject anatomical variation to mitigate its effect on the ECG inverse problem \cite{journal/TBME/gyawali2021}.
Besides the heart, the torso geometry accuracy can also subtly influence the inverse solution, as the EP measurement is generally performed on the body surface \cite{conf/CinC/zemzemi2015}.
There exist several studies performed UQ to investigate the effects of cardiac position inside the torso and torso conductivity on the inverse inference \cite{conf/CiC/bergquist2022,conf/FIMH/zemzemi2015,journal/PM/bergquist2023}.
To address uncertainties, advanced regularization techniques that encompass prior knowledge, noise reduction strategies, and advanced data assimilation methods show promise \cite{journal/TMI/rodrigo2017,journal/JNMBE/gander2021}. 

Model assumptions are necessary simplifications or idealizations of reality that are made to facilitate computational modeling. 
In the ECG inverse inference, assumptions are often made about the homogeneity of tissue properties, the behavior of electrical signals within the heart, and the relationship between measured ECG signals and underlying electrical activity \cite{conf/FIMH/tate2021,journal/A/zhao2020,journal/FP/van2021}.
While model assumptions are essential for simplifying complex systems, they can also introduce uncertainties and errors into the modeling process. 
For example, assuming uniform tissue properties throughout the heart and torso may lead to inaccuracies in the predicted electrical activity, especially in regions with significant heterogeneity \cite{conf/CinC/zemzemi2015,journal/CMPB/yadan2023}. 
Similarly, assuming specific boundary conditions or physiological behaviors may not fully capture the complexity of real-world scenarios, leading to discrepancies between model predictions and observed data.
ECG inverse inference also depends heavily on the assumptions on the initial distribution of heart electrical potential \cite{journal/JP/mirams2016}.
Determining the initial potential distribution at the onset of the cardiac cycle is particularly challenging due to the difficulty in accurately pinpointing the anatomical site of the earliest excitation in the heart, often associated with the intricate Purkinje networks \cite{journal/TASE/xie2023}. 
Therefore, sophisticated modeling strategies are required to adequately account for the influence of these networks on ECG inverse inference \cite{journal/JCP/palamara2015,journal/MedIA/camps2024}. 
Once uncertainties are quantified, strategies can be employed to mitigate their impact on model predictions.
The process of balancing model assumptions and uncertainty is often iterative, requiring continuous refinement and validation of models against experimental or clinical data. 
This iterative refinement process helps researchers to build more robust and reliable models over time.

\subsection{Surrogate Model for Efficient ECG Simulation in Cardiac Digital Twin Development} \label{discussion:surrogate}

The functional twinning of cardiac EP often requires extensive simulations to achieve personalized parameter optimization.
Traditional methods for simulating ECG signals involve complex biophysical models that account for intricate cardiac electrical and anatomical properties. 
While these models offer high fidelity, they often require substantial computational resources and time \cite{journal/TBME/potse2006}. 
Fast implementations of these models exist. 
For example, Sachetto \textit{et al.} \cite{journal/JNMBE/sachetto2018} presents a GPU-based implementation of the monodomain model, which is capable of speeding up simulations by a factor up to 498 when compared to a serial implementation of the model. 
They reported simulation times of 32 min per heartbeat. 
Other studies have proposed alternative models that can yield similar simulation outputs at a fraction of the state-of-the-art monodomain models. 
For example, Wallman \textit{et al.} \cite{journal/MedIA/wallman2014} proposed a graph-based Eikonal model that can simulate the activation sequence in the heart in less than 1.6 s \cite{journal/MedIA/camps2021}. 
Similarly, Neic \textit{et al.} \cite{journal/JCP/neic2017} introduced a reaction-Eikonal model, later implemented with a full pseudo-bidomain approach by Gillette \textit{et al.} \cite{journal/MedIA/gillette2021}, which reported a computational time of 213 s per heartbeat. 
A simplified version using a lead field approach, without repolarization, can potentially be 10 times faster.
These studies demonstrate a range of efficient ECG simulation tools, though reported times can vary significantly depending on model complexity (e.g., isotropic vs. anisotropic models), mesh resolution, and the hardware used in each approach \cite{journal/FiP/pezzuto2017, journal/MedIA/gillette2021}.
Nevertheless, in the context of generating digital twins, the number of simulations required per patient can easily be in the range of hundreds of thousands \cite{journal/MedIA/camps2021,journal/MedIA/camps2024}. 
Thus, even the fastest EP models available present a significant computational burden when applied at scale for digital twin generation \cite{journal/MedIA/gillette2021,journal/MedIA/camps2024}.

The emergence of neural surrogate models addresses these challenges by offering computationally efficient alternatives that balance accuracy with practicality \cite{journal/FCM/herrero2022,conf/MICCAI/jiang2022,conf/FIMH/bertrand2023}. 
These surrogate models leverage machine learning techniques to learn the intricate mapping between the physiological parameters (e.g., EASs) and expected behavior (e.g., cardiac activation sequences) \cite{conf/FIMH/bertrand2023}. 
For example, Bertrand \textit{et al.} \cite{conf/FIMH/bertrand2023} was able to produce simulations of the activation sequence based on U-Net in 0.04 s, which was a 500-factor speed-up compared to a serial implementation of the graph-based Eikonal.
Their surrogate EP model was generic and simplified (CV was fixed) and required an additional 5h 32 min of training time, but once trained, it can be deployed with high efficiency.
Instead, Jiang \textit{et al.} \cite{conf/MICCAI/jiang2022} developed a personalized neural surrogate model using meta-learning, reducing simulation time to 0.24 s and personalization time to 0.032 s, with an overall training time of about 8.5 min, significantly faster than the conventional 5 min AP simulation requiring 100 simulation calls.
By circumventing the need for exhaustive simulations, surrogate models provide an avenue for rapid and on-demand generation of cardiac functional data (e.g., ECG signals). 
This is particularly valuable in scenarios where real-time simulations or large-scale analyses are necessary. 
However, challenges such as model interpretability, generalization across diverse populations, and the need for accurate training data warrant careful consideration. 
Furthermore, a thorough understanding of the trade-off between computational efficiency and simulation accuracy is vital when integrating surrogate models into clinical applications. 
As surrogate models continue to evolve, they hold the potential to revolutionize CDTs by enabling researchers to efficiently explore the parameter space and improve our understanding of complex cardiac phenomena.

\subsection{Multimodal Representation Learning Towards Solving the ECG Inverse Problem} \label{discussion:multimodal}

The ECG inverse inference involves the fusion of information from multiple modalities, such as ECG signals, anatomical images (e.g., CT/ MRI scans), and EP parameters.
Multimodal representation learning, a technique widely used in the computer vision field, holds significant potential for ECG inverse inference by enabling the seamless integration of diverse data sources \cite{journal/STSP/zhang2020}.
Deep neural networks have proven to be a highly effective strategy for ECG interpretation and analysis via spatiotemporal representation learning \cite{journal/CBM/hong2020}. 
However, most works only considered ECG data for analysis and only a few works employed joint analysis with other data sources \cite{journal/EP/bacoyannis2021,conf/STACOM/li2022,journal/FiP/beetz2022,journal/TMI/li2024}.
Among these multimodal works, VAE has been widely employed to learn a latent representation of ECG and geometry \cite{conf/ISBI/beetz2022,journal/TMI/li2024}, along with incorporating uncertainty modeling for inverse inference \cite{journal/EP/bacoyannis2021}.
However, these studies merely concatenated the acquired features or utilized a shared representation, lacking a thorough explanation of how these features align and contribute to holistic data comprehension.
In contrast, physics-informed models can integrate domain-specific knowledge and principles derived from the underlying physics of the cardiac system \cite{journal/TMI/jiang2024}.
By explicitly encoding the physical interactions and dependencies between ECG and geometry, these models could offer a more transparent and interpretable way to capture the intricate interplay between these data modalities. 
Moreover, they could potentially enhance the generalization capabilities of the learned representations, making them more robust in scenarios with limited or noisy data.
Nevertheless, the integration of physics-based constraints into machine learning models introduces additional challenges, such as learning geometrical invariance, as well as processing model complexity and the need for accurate domain knowledge. 
Balancing the complexity of the physical model with the capacity of the machine learning algorithm while also ensuring the interpretability of the resulting features will require careful consideration and innovative methodologies.

\section{Conclusion} \label{conclusion}

CDTs are pivotal in advancing precision medicine and improving patient outcomes in cardiovascular care by simulating and predicting personalized responses to interventions. 
However, current models encounter a computational bottleneck in accurately calibrating CDTs to individual physiology through ECG inverse inference, particularly in estimating EP parameters within 3D heart models.
This review comprehensively outlines recent progress in ECG inverse inference, encompassing algorithms, public datasets, evaluation, clinical applications, and future perspectives. 
While machine learning, especially deep learning, holds promise in tackling this complex and ill-posed problem, significant challenges persist, alongside promising opportunities for further advancement and innovation.
In the future, as these techniques gain increasing popularity and generate massive cohorts of digital twins, they are poised to revolutionize cardiovascular care through in-silico trials and personalized interventions. 
With the growing availability of big data, we can envision a future where everyone possesses a digital twin, significantly reducing the reliance on animal models and enabling real-time monitoring and precision treatment in clinical settings.

\bibliographystyle{ieeetr}
\bibliography{A_refs}

\end{document}